\documentclass{article}

\usepackage{microtype}
\usepackage{graphicx}
\usepackage{subfigure}
\usepackage{booktabs} 

\usepackage{amsmath,amssymb,enumitem,multirow,mathtools}

\newtheorem{theorem}{Theorem}

\newtheorem{lemma}{Lemma}

\usepackage{hyperref}

\usepackage[accepted]{icml2019}

\icmltitlerunning{Robust Inference via Generative Classifiers for Handling Noisy Labels}

\begin{document}

\twocolumn[
\icmltitle{Robust Inference via Generative Classifiers for Handling Noisy Labels}




\begin{icmlauthorlist}
\icmlauthor{Kimin Lee}{ka}
\icmlauthor{Sukmin Yun}{ka}
\icmlauthor{Kibok Lee}{umich}
\icmlauthor{Honglak Lee}{goo,umich}
\icmlauthor{Bo Li}{uiuc}
\icmlauthor{Jinwoo Shin}{ka,aitrics}
\end{icmlauthorlist}

\icmlaffiliation{ka}{KAIST}
\icmlaffiliation{goo}{Google Brain}
\icmlaffiliation{uiuc}{University of Illinois at Urbana Champaign}
\icmlaffiliation{umich}{University of Michingan Ann Arbor}
\icmlaffiliation{aitrics}{AItrics}

\icmlcorrespondingauthor{Kimin Lee}{kiminlee@kaist.ac.kr}

\icmlkeywords{Machine Learning, ICML}

\vskip 0.2in
]



\printAffiliationsAndNotice{} 

\begin{abstract}
\vspace{-0.07in}
Large-scale datasets may contain significant proportions of noisy (incorrect) class labels, and it is well-known that modern deep neural networks (DNNs) poorly generalize from such noisy training datasets. 
To mitigate the issue,
we propose a novel inference method, termed \emph{Robust Generative classifier (RoG)}, applicable
to any discriminative (e.g., softmax) neural classifier pre-trained on noisy datasets.
In particular, we induce a generative classifier on top of hidden feature spaces of the pre-trained DNNs,
for obtaining a more robust decision boundary.
By estimating the parameters of generative classifier using the minimum covariance determinant estimator, we significantly improve the classification accuracy with neither re-training of the deep model nor changing its architectures.
With the assumption of Gaussian distribution for features,
we prove that RoG generalizes better than baselines under noisy labels. 
Finally, we propose the ensemble version of RoG to improve its performance by investigating the layer-wise characteristics of DNNs. Our extensive experimental results demonstrate the superiority of RoG given different learning models optimized by several training techniques to handle diverse scenarios of noisy labels.
\end{abstract}

\begin{figure*}[t] \centering
\vspace{-0.1in}
\subfigure[Test set accuracy comparison]
{
\includegraphics[width=0.3\textwidth]{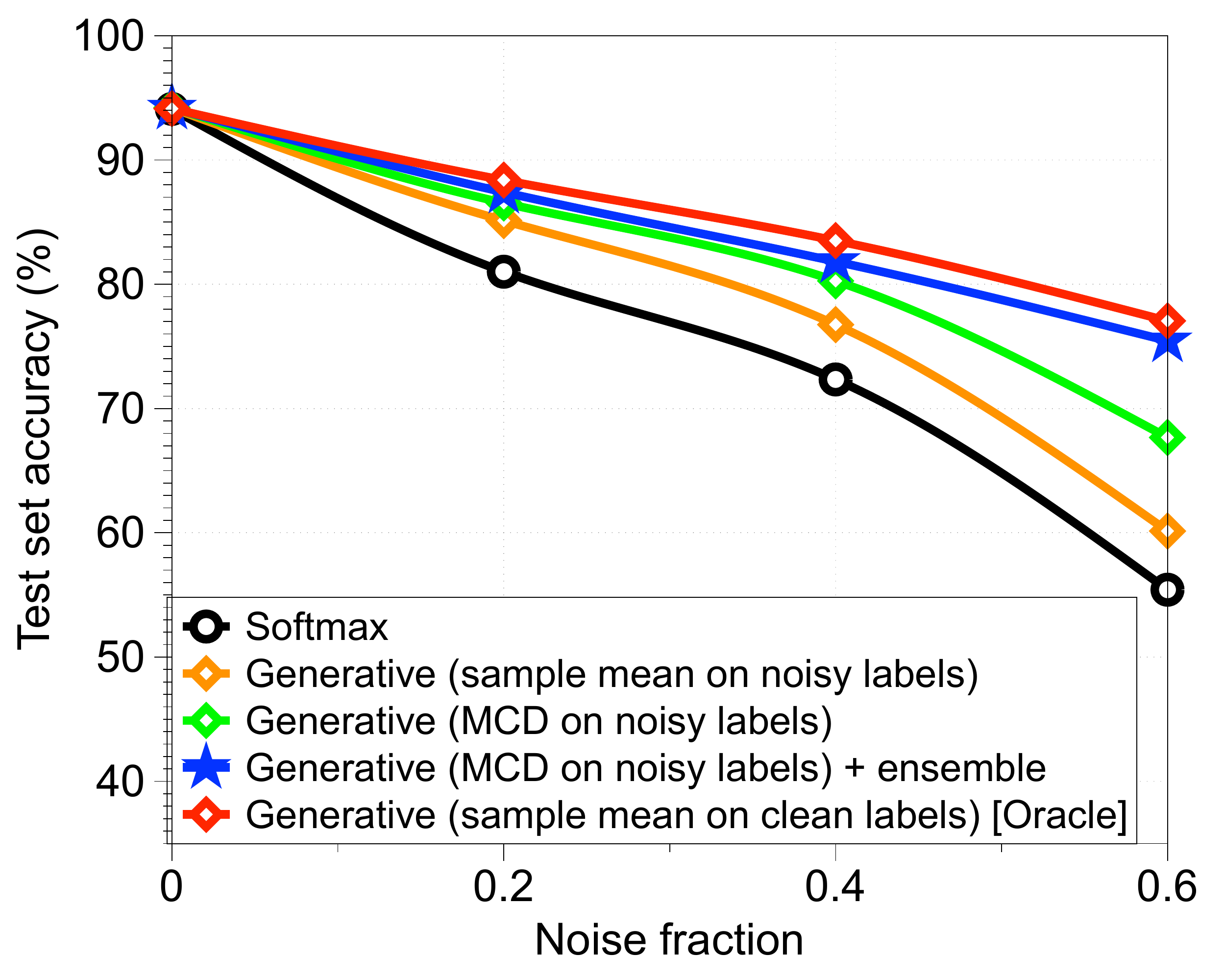} \label{fig:acc_noise}}
\,
\subfigure[Penultimate features by t-SNE]
{
\includegraphics[width=0.3\textwidth]{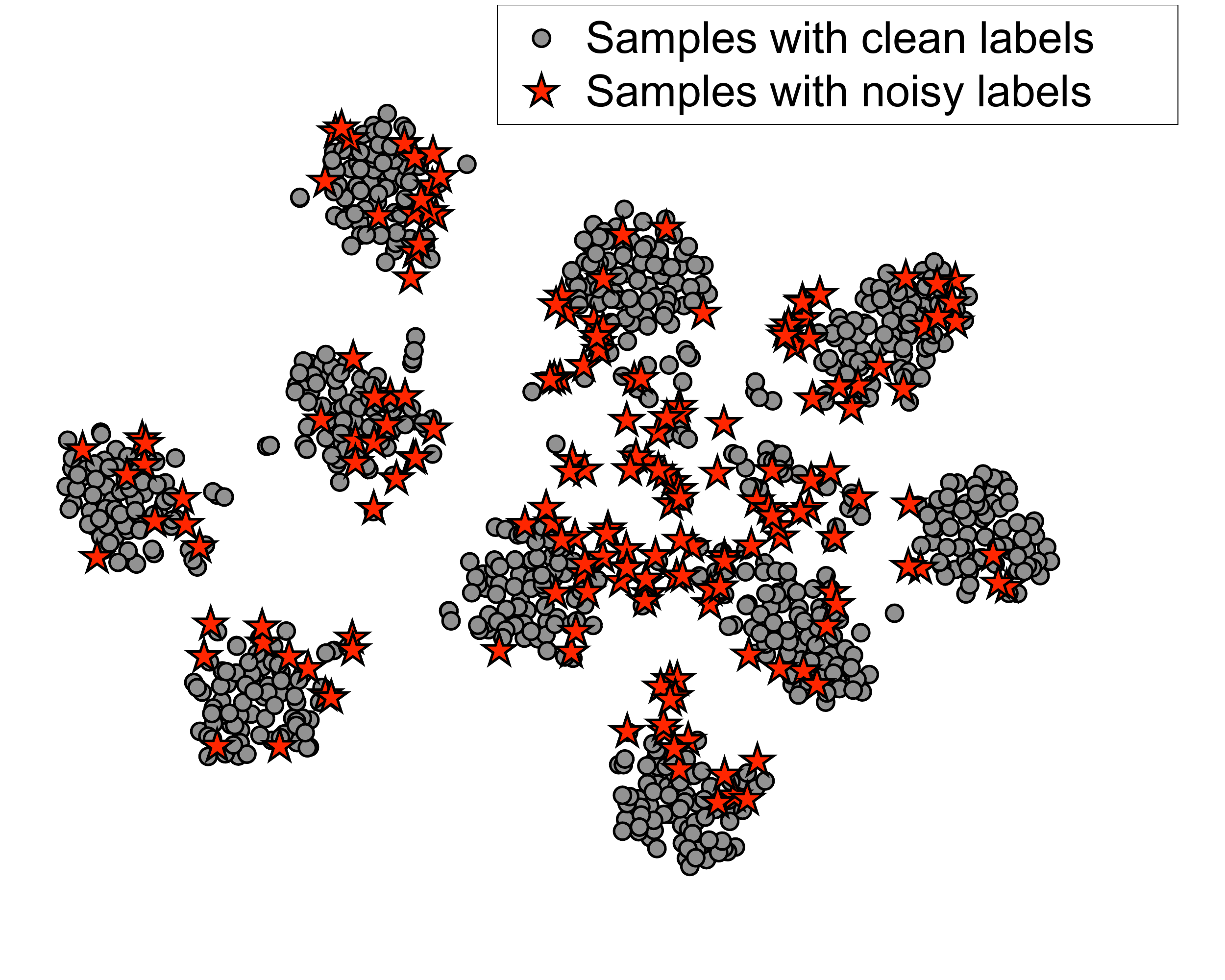} \label{fig:tsne_train}}
\,
\subfigure[An illustration of the MCD estimator]
{
\includegraphics[width=0.3\textwidth]{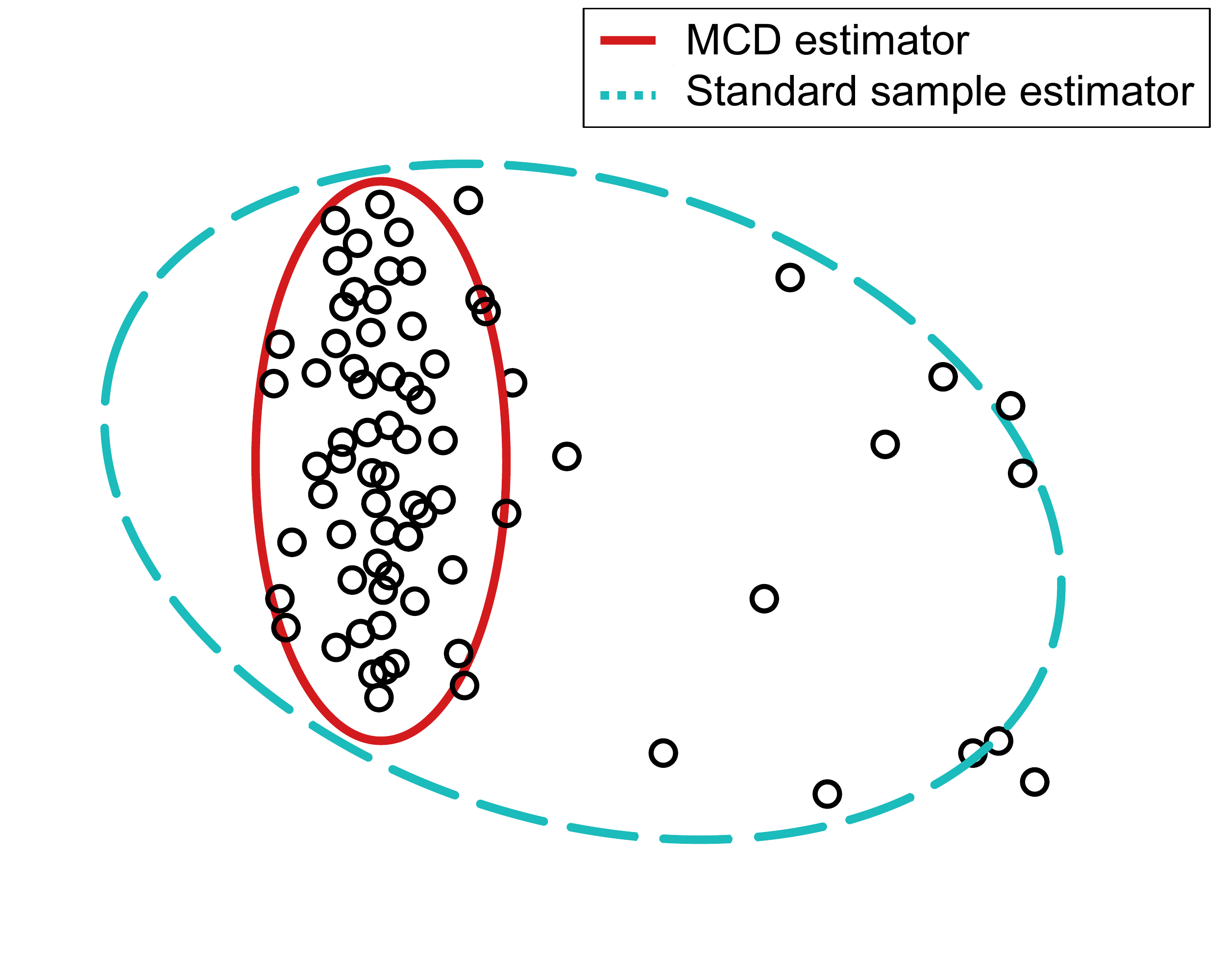} \label{fig:mcd}}
\vspace*{-0.05in}
\caption{Experimental results under DenseNet-100 and CIFAR-10 with uniform noise, i.e., the labels of a given proportion of training samples are flipped to other labels uniformly at random.
(a) Test set accuracy of softmax and generative classifiers with various parameter estimations.
(b) Visualization of features on the penultimate layer using t-SNE from training samples when the noise fraction is 20\%. (c) An illustration of the MCD estimator: it is more robust against outliers by finding a subset with minimum covariance determinant.}
\label{fig:motivation}
\vspace*{-0.15in}
\end{figure*}

\vspace*{-0.3in}
\section{Introduction}
\label{sec:intro}
\vspace{-0.05in}

Deep neural networks (DNNs) tend to generalize well when they are trained on large-scale datasets with ground-truth label annotations.
For example, DNNs have achieved state-of-the-art performance on many classification tasks, e.g., image classification \citep{he2016deep}, object detection \citep{girshick2015fast}, and speech recognition \citep{amodei2016deep}.
However, as the scale of training dataset increases, it becomes infeasible to obtain all ground-truth class labels from domain experts.
A common practice is collecting the class labels from data mining on social media \citep{mahajan2018exploring} or web data \citep{krause2016unreasonable}. 
Machine-generated labels are often used; e.g., the Open Images Dataset V4 contains
such 70 million labels for training images \citep{kuznetsova2018open}.
However, they may contain incorrect labels, and
recent studies have shown that modern deep architectures may generalize poorly from the noisy
datasets 
\citep{zhang2016understanding} (e.g., 
see the black line of Figure~\ref{fig:acc_noise}).

To address the poor generalization issue of DNNs with noisy labels, many training strategies
have been investigated 
\citep{reed2014training,patrini2017making,ma2018dimensionality,han2018co,hendrycks2018using,goldberger2016training,jiang2017mentornet,ren2018learning,zhang2018generalized,malach2017decoupling,han2018pumpout}. 
However, using such training methods may incurs expensive back-and-forth costs (e.g., additional time and hyperparameter tuning) and suffer from the reproducibility issue. 
This motivates our approach of developing a more plausible inference method applicable to any pre-trained deep model.
Hence, our direction is complementary to the prior works: one can combine ours and a prior training method for the best performance (see Tables~\ref{tbl:comparison_with_SOTA}, \ref{tbl:main_on_co-teaching}, \& \ref{tbl:comparison_with_SOTA_nlp} in Section \ref{sec:experiments}).

The key contribution of our work is to develop such an inference method, \emph{Robust Generative classifier (RoG)}, which is applicable to any discriminative (e.g., softmax) neural classifier pre-trained on noisy datasets (without re-training). 
Our main idea is inducing a better posterior distribution from the pre-trained (noisy, though) feature representation by utilizing a robust
generative classifier. 
Here, our belief is that the softmax
DNNs can learn meaningful feature patterns shared by multiple training examples even under datasets with noisy labels, e.g., see \citep{arpit2017closer}.

To motivate our approach, 
we first observe that training samples with noisy labels (red circles) are distributed like outliers when their hidden features are projected in a 2-dimensional space using t-SNE \citep{maaten2008visualizing} (see Figure~\ref{fig:tsne_train}).
In other words, this phenomena implies that DNN representations even when trained with noisy labels may still exhibit \emph{clustering properties} (i.e., the DNN learns embedding that tend to group clean examples of the same class into the clusters while pushing away the examples with corrupt labels outside these clusters). 
The observation inspires us to induce a generative classifier on the pre-trained hidden features since it can model joint data distributions $P(x,y)$ for input $x$ and its label $y$ for outlier detection and thus produce robust posterior $P(y\,|\,x)$ for prediction. 
Here, one may suggest to train a deep generative classifier from scratch. However,
such a fully generative approach is expensive and has been not popular for recent state-of-art classification.
We instead post-process a light generative classifier only
for inference. 


In particular, we propose to induce the generative classifier under linear discriminant analysis (LDA) assumption and
choose its parameters by the minimum covariance determinant (MCD) \citep{rousseeuw1984least} estimator which calculates more robust parameters. 
We provide a theoretical support on the generalization property \citep{durrant2010compressed} of RoG based on MCD:
it has the smaller errors on parameter estimations provably under some Gaussian assumptions.
To improve RoG further, 
we observe that RoG built from low-level features can be often more effective since DNNs tend to have similar hidden features, regardless of whether they are trained with clean or noisy labels at early layers \citep{arpit2017closer, morcos2018insights}.
Under the observations, we finally propose an ensemble version of 
RoG to incorporate all effects of low and high layers.

We demonstrate the effectiveness of RoG using modern neural architectures 
on image classification and natural language processing tasks.
In all tested cases,
our methods (e.g., see green and blue lines in Figure~\ref{fig:acc_noise}) significantly outperform the softmax classifier, although they use the same feature representations trained by the noisy dataset.
In particular, we show that RoG can be used to further improve various prior training methods
\citep{reed2014training, patrini2017making, ma2018dimensionality, han2018co, hendrycks2018using} which are specialized to handle the noisy environment.
For example, we improve the test accuracy of
the state-of-the-art training method \cite{han2018co} on CIFAR-100 dataset with 45\% noisy labels from 33.34\% to 43.02\%.
Finally, RoG is shown to be working properly against more semantic noisy labels (generated from a machine labeler) and open-set noisy labels \citep{wang2018iterative}.
\section{Related work}
\vspace{-0.05in}
One of major directions for handling noisy labels is utilizing an estimated/corrected labels during training: \citet{reed2014training} proposed a bootstrapping method which trains deep models with new labels generated by a convex combination of the raw (noisy) labels and their predictions, and \citet{ma2018dimensionality} improved the bootstrapping method by utilizing the dimensionality of subspaces during training.
\citet{patrini2017making} modified the loss and posterior distribution to eliminate the influence of noisy labels, and \citet{hendrycks2018using} improved such a loss correction method by utilizing the information from data with true class labels. 
Another promising direction has focused on training on selected (cleaner) samples: \citet{jiang2017mentornet} introduced a meta-learning model, called MentorNet, and \citet{han2018pumpout} proposed a meta approach which can improve MentorNet. \citet{ren2018learning} adaptively assigned weights to training samples based on their gradient directions.
\citet{malach2017decoupling} and \citet{han2018co} proposed the selection methods based on an ensemble of deep models.
Compared the above training methods, adopting the above training methods might incur expensive back-and-forth costs. On the other hand,
our generative inference method is very cheap and enjoys an orthogonal usage, i.e., ours can be easily applied to improve 
any of them.

Inducing a generative classifier (e.g., a mixture of Gaussian) on pre-trained deep models also has been investigated for various purposes:
\citet{hermansky2000tandem} propose Tandem approaches which induce a generative model on top of hidden features for speech recognition.
More recently, by inducing the generative model, \citet{lee2018simple} introduce the Mahalanobis distance-based confidence score for novelty detection.
However, their methods use naive parameter estimation under assuming perfect clean training labels,
which 
should be highly influenced by outliers.
We overcome the issue by using the MCD estimator. 

\section{Robust Inference via Generative Classifiers}
\label{sec:inference}

In this section, we propose a novel inference method which obtains a robust posterior distribution from any softmax neural classifier pre-trained on datasets with noisy labels.
Our idea is inducing the generative classifier given hidden features of the deep model.
We show the robustness of our method in terms of high breakdown points \citep{hampel1971general}, and generalization error \citep{durrant2010compressed}.
We also investigate the layer-wise characteristics of generative classifiers, and introduce an ensemble of them to improve its performance. 


\subsection{Generative Classifier and MCD Estimator} 
\label{section:thm}

Let $\mathbf{x}$ be an input and $y \in  \{1,\cdots,C\}$ be its class label.
Without loss of generality, 
suppose that
a pre-trained softmax neural classifier is given:
$P\left(y=c|\mathbf{x}\right)
= \frac{  \exp \left( \mathbf{w}_c^\top f \left( \mathbf{x} \right) + b_c \right) }{\sum_{c^\prime} \exp \left( \mathbf{w}_{c^\prime}^\top f \left( \mathbf{x} \right)  + b_{c^\prime}\right) },$
where $\mathbf{w}_c$ and $b_c$ are the weight and the bias of the softmax classifier for class $c$, and $f(\cdot)\in \mathbb{R}^d$ denotes the output of the penultimate layer of DNNs.
Then, without any modification on the pre-trained softmax neural classifier,
we induce a generative classifier by assuming the class-conditional distribution follows the multivariate Gaussian distribution. 
In particular, 
we define $C$ Gaussian distributions with a tied covariance $\mathbf{\Sigma}$, i.e.,
linear discriminant analysis (LDA) \citep{fisher1936use}, and a Bernoulli distribution for the class prior: 
$P\left( f(\mathbf{x})|y=c\right) = \mathcal{N} \left(f(\mathbf{x})| \mathbf{\mu}_c, \mathbf{\Sigma} \right), P\left(y=c\right) = \beta_c,$
where $\mathbf{\mu}_c$ is the mean of multivariate Gaussian distribution and $\beta_c$ is the normalized prior for class $c$.
We provide an analytic justification on the LDA (i.e., tied covariance) assumption in the supplementary material.
Then, based on the Bayesian rule,
we induce a new posterior different from the softmax one as follows:
\begin{align*}
\vspace{-0.05in}
& P\left(y=c|f(\mathbf{x})\right) = \frac{P\left( y = c \right) P\left(f(\mathbf{x})| y =c \right) }{\sum \limits_{c^\prime}P\left( y= c^\prime \right) P\left(f(\mathbf{x})| y= c^\prime \right)} \\
&= \frac{ \exp \left( \mathbf{\mu}_c^\top \mathbf{\Sigma}^{-1} f(\mathbf{x}) -\frac{1}{2} \mathbf{\mu}_c^\top \mathbf{\Sigma}^{-1} \mathbf{\mu}_c +\log \beta_c \right) }{\sum \limits_{c^\prime} \exp \left( \mathbf{\mu}_{c^\prime}^\top \mathbf{\Sigma}^{-1} f(\mathbf{x}) -\frac{1}{2} \mathbf{\mu}_{c^\prime}^\top \mathbf{\Sigma}^{-1} \mathbf{\mu}_{c^\prime} +\log \beta_{c^\prime} \right)}.
\vspace{-0.05in}
\end{align*}
To estimate the parameters of the generative classifier,
one can compute the sample class mean and covariance of training samples $\mathcal{X}_N = \{(\mathbf{x}_1,y_1),\ldots, (\mathbf{x}_N,y_N)\}$:
\begin{align} \label{eq:sample_estimator}
&{\bar \mu}_c  =  \sum_{i:y_i=c}  \frac{f(\mathbf{x}_i)}{N_c}, \quad {\bar \beta}_c  = \frac{N_c}{N}, \notag \\
&\mathbf{\bar \Sigma} = \sum_c \sum_{i:y_i=c}  \frac{\left(f(\mathbf{x}_i) - \bar \mu_c\right) \left(f(\mathbf{x}_i) - \bar \mu_c\right)^\top}{N},
\end{align}
where $N_c$ is the number of samples labeled to be class $c$.

However, one can expect that the naive sample estimator (\ref{eq:sample_estimator}) can be highly influenced by outliers (i.e., training samples with noisy labels).
In order to improve the robustness,
we propose the so-called {\em Robust Generative classifier (RoG)}, which utilizes the minimum covariance determinant (MCD) estimator \citep{rousseeuw1999fast} to estimate its parameters.
For each class $c$, the main idea of MCD is finding a subset $\mathcal{X}_{K_c}$ for which the determinant of the corresponding sample covariance is minimized:
\begin{align}
  \min \limits_{\mathcal{X}_{K_c} \subset \mathcal{X}_{N_c}} \text{det} \left( {\mathbf {\widehat \Sigma}}_c \right)\quad \text{subject to} ~ |\mathcal{X}_{K_c}| = K_c,\label{eq:mcdopt}
\end{align} 
where $\mathcal{X}_{N_c}$ is the set of training samples labeled to be class $c$, ${\mathbf {\widehat \Sigma}}_c$ is the sample covariance of $\mathcal{X}_{K_c}$ and $0<K_c<N_c$ is a hyperparameter.
Then, only using the samples in $\bigcup_c~ \mathcal{X}_{K_c}$,
it estimates the parameters, i.e., ${\widehat \mu}_c, \mathbf{\widehat \Sigma}, {\widehat \beta}_c$, of the generative classifier, by following (\ref{eq:sample_estimator}).
{Such a new estimator can be more robust by removing the outliers which might be widely scattered in datasets (see Figure~\ref{fig:mcd}).}

The robustness of MCD estimator has been justified in the literature: it is known to have near-optimal breakdown points \citep{hampel1971general}, i.e., the smallest fraction of data points that need to be replaced by arbitrary values (i.e., outliers) to fool the estimator completely.
Formally,
denote $\mathcal{Y}_{M}$ as a set obtained by replacing $M$ data points of set $\mathcal{Y}$ by some arbitrary values. Then, for a multivariate mean estimator $\mu=\mu(\mathcal Y)$ from $\mathcal Y$, the breakdown point is defined as follows (see the supplementary material for more detailed explanations including the breakdown point of covariance estimator):
\begin{align*}
    &\varepsilon^* (\mu, \mathcal{Y}) \\
    &= \frac{1}{|\mathcal Y|} \min \left\{ M \in \left[|\mathcal Y|\right] : \sup_{\mathcal{Y}_{M}} \left\| \mu(\mathcal{Y}) -\mu(\mathcal{Y}_{M}) \right\| = \infty \right\},
\end{align*}
where the set $\{1,\ldots, n \}$ is denoted by $[n]$ for positive integer $n$.
While the breakdown point of the naive sample estimator is 0\%, 
the MCD estimator for the generative classifier under LDA assumption is known to
attain near optimal breakdown value of $\min_c \frac{\lfloor (N_c - d +1) / 2 \rfloor}{N_c} \approx 50\%$ \citep{lopuhaa1991breakdown}.
Inspired by this fact,
we choose the default value of {$K_c$ in (\ref{eq:mcdopt})
by $\lfloor (N_c+d+1)/2 \rfloor$.}

We also establish the following theoretical support that the MCD-based generative classifier (i.e., RoG) can have smaller errors on parameter estimations, compared to the naive sample estimator, under some assumptions for its analytic tractability. 
\begin{theorem} \label{th:main}
Assume the followings:
\begin{itemize}
    \item[($\mathcal A1$)] {For (clean) sample $\mathbf x$ of correct label, the class-conditional distribution of hidden feature $f(\mathbf x)$ of DNNs has mean $\mathbf{\mu}_c$ and tied covariance matrix $\sigma^2 {\mathbf I}$.
    For (outlier) sample $\mathbf x$ of incorrect label, the distribution of hidden feature has mean $\mu_{\tt out}$ and covariance matrix $\sigma_{\tt out}^2 {\mathbf I}$, where ${\mathbf I} \in \mathbb R^{d\times d}$ is the identity matrix.}
    \item[($\mathcal A2$)] All classes have the same number of samples (i.e., $N_c=\frac{N}C$), the same fraction $\delta_{\tt out}<1$ of outliers, and the sample fraction $\delta_{\tt mcd}=\frac{K_c}{N_c}<1$ of samples selected by MCD estimator.
    \item[($\mathcal A3$)] The outliers are widely scattered such that {$\sigma^2 < \sigma_{\tt out}^2$.}
    \item[($\mathcal A4$)] The number of outliers is not too large such that $\delta_{\tt out}<1-\delta_{\tt mcd}$ and $\delta_{\tt mcd}>\frac{d}{N_c}$.
\end{itemize}
Let $\widehat \mu, \mathbf{\widehat \Sigma}$ and $\bar \mu, \mathbf{\bar \Sigma}$ be the {outputs} of the MCD and sample estimators, respectively. Then, for all $c,~c^\prime$, {$\widehat \mu, \bar \mu, \mathbf{\widehat \Sigma}, \mathbf{\bar \Sigma}$ converge almost surely to their expectation as $N\rightarrow\infty$}, and it holds that 
\begin{align}
    &\|\mu_c - {\widehat \mu}_c\|_1\overset{\tt a.s.}{\rightarrow}\lim_{N\rightarrow\infty}\|\mu_c - {\widehat \mu}_c\|_1 = 0, \notag \\
    & \|\mu_c - {\Bar \mu}_c\|_1\overset{\tt a.s.}{\rightarrow}\lim_{N\rightarrow\infty}\|\mu_c - {\Bar \mu}_c\|_1 = \delta_{\tt out}\|\mu_c\|_1, \label{ineq1:mahalanobis} \\ 
    &\frac{\phi({\mathbf {\widehat \Sigma}})\|\widehat{\mu}_c - \widehat{\mu}_{c^\prime}\|_2}{\phi({\mathbf {\Bar \Sigma}})\|\Bar{\mu}_c - \Bar{\mu}_{c^\prime}\|_2} \overset{\tt a.s.}{\rightarrow}
    \lim_{N\rightarrow\infty}\frac{\phi({\mathbf {\widehat \Sigma}})\|\widehat{\mu}_c - \widehat{\mu}_{c^\prime}\|_2}{\phi({\mathbf {\Bar \Sigma}})\|\Bar{\mu}_c - \Bar{\mu}_{c^\prime}\|_2} \notag \\
    & \; \;\qquad \quad \quad \quad \quad \quad = \; \lim_{N\rightarrow\infty} \frac{1}{\left(1-\delta_{\tt out}\right)^2 \phi({\mathbf {\Bar \Sigma}})} \geq 1, \label{ineq2:mahalanobis}
\end{align}
where $\phi({\mathbf {\widehat \Sigma}}) = {4{\|\mathbf {\widehat \Sigma}^{-1}\|_2}{\|\mathbf {\widehat \Sigma}\|_2}}
{\left(1+{\|\mathbf {\widehat \Sigma}^{-1}\|_2}{\|\mathbf {\widehat \Sigma}\|_2}\right)^{-2}}$. 
\end{theorem}
The proof of the above theorem is given in the supplementary material, 
where it is built upon the fact that the determinants can be expressed as the $d$-th degree polynomial of outlier ratio under the assumptions.
We note that one might enforce the assumptions of the diagonal covariance matrices in $\mathcal A1$ to hold under an affine translation of hidden features.
In addition, the assumption in $\mathcal A4$ holds when $N_c$ is large enough. 
Nevertheless, we think most assumptions of Theorem \ref{th:main} are not necessary
to claim the superiority of RoG 
and it is an interesting future direction to explore to relax them.

The generalization error bound of a generative classifier under the assumption that the
class-conditional Gaussian distributions of features
is known to be bounded as follows \citep{durrant2010compressed}:
\begin{align*}
&P_{f(\mathbf{x})} \left( y^* \neq \arg\max_y P_{{\widehat \mu}_c, {\widehat \Sigma}}(y|f(\mathbf{x})) \right) \\ 
    &\leq \sum_c  \sum_{c^\prime \neq c} \exp \left(-\frac{\|{\widehat \mu}_c - {\widehat \mu}_{c^\prime}\|_2}{8\sigma^2}  \cdot \phi(\mathbf {\widehat \Sigma})\right) + D\|\mu_c - {\widehat \mu}_c\|_1,
\end{align*}
for some constant $D>0$. Therefore, (\ref{ineq1:mahalanobis}) and (\ref{ineq2:mahalanobis}) together imply that utilizing the MCD estimator provides a better generalization bound, compared to the sample estimator.

\begin{algorithm}[t]
\caption{\citep{rousseeuw1999fast}  Approximating MCD for a single Gaussian.} \label{alg:MCD_single}
\begin{algorithmic}[1]
\STATE {\bf Input:} $\mathcal{X}_{N_c}=\{\mathbf{x}_i : i =1,\cdots,N_c \}$ and the maximum number of iterations $I_{\max}$. \vspace{0.05in}
\hrule
\vspace{0.05in}
\STATE Uniformly sample initial subset $\mathcal{X}_{K_c} \subset \mathcal{X}_{N_c}$, where $|\mathcal{X}_{K_c}| = \lfloor (N_c+d+1)/2 \rfloor$.
\STATE Compute a mean ${\widehat \mu}_c  =  \frac{1}{|\mathcal{X}_{K_c}|} \sum\limits_{\mathbf{x} \in \mathcal{X}_{K_c}} f(\mathbf{x})$, and covariance $\mathbf{\widehat \Sigma}_c = \frac{1}{|\mathcal{X}_{K_c}|} \sum \limits_{\mathbf{x} \in \mathcal{X}_{K_c}} \left(f(\mathbf{x}) - \widehat \mu_c \right) \left(f(\mathbf{x}) - \widehat \mu_c \right)^\top.$
\FOR{$i=1$ {\bfseries to} $I_{\max}$} 
\STATE Compute the Mahalanobis distance for all $\mathbf{x} \in \mathcal{X}_{N_c}$: \\
$\alpha (\mathbf{x}) = \left(f(\mathbf{x}) - \mathbf{\widehat \mu_c} \right)^\top \mathbf{\widehat \Sigma}_c^{-1} \left(f(\mathbf{x}) - \mathbf{\widehat \mu_c} \right).$
\STATE Update $\mathcal{X}_{K_c}$ such that it includes $\lfloor (N_c+d+1)/2 \rfloor$ samples with smallest distance $\alpha (\mathbf{x})$.
\STATE Compute sample mean and covariance, i.e., $\widehat \mu_c, \mathbf{\widehat \Sigma}_c$, using new subset $\mathcal{X}_{K_c}$.
\STATE Exit the loop if the determinant of covariance matrix is not decreasing anymore.
\ENDFOR
\STATE Return $\widehat \mu_c$ and $\mathbf{\widehat \Sigma}_c$
\end{algorithmic}
\end{algorithm}

\begin{figure*} [t] \centering
\vspace{-0.1in}
\subfigure[Model comparison]
{
\includegraphics[width=0.29\textwidth]{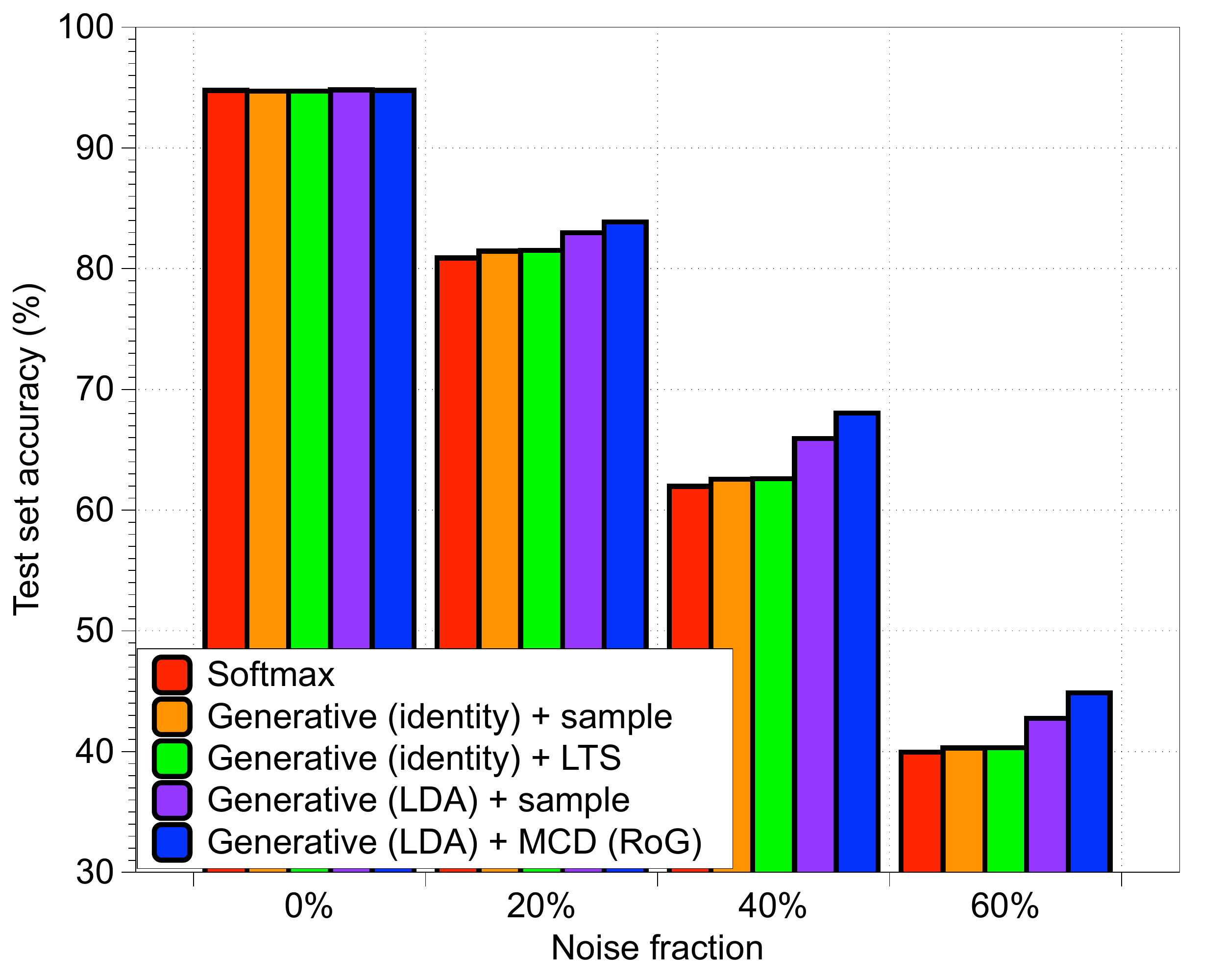} \label{fig:l_acc_noise}} 
\,
\subfigure[Accuracy of the MCD estimator]
{
\includegraphics[width=0.29\textwidth]{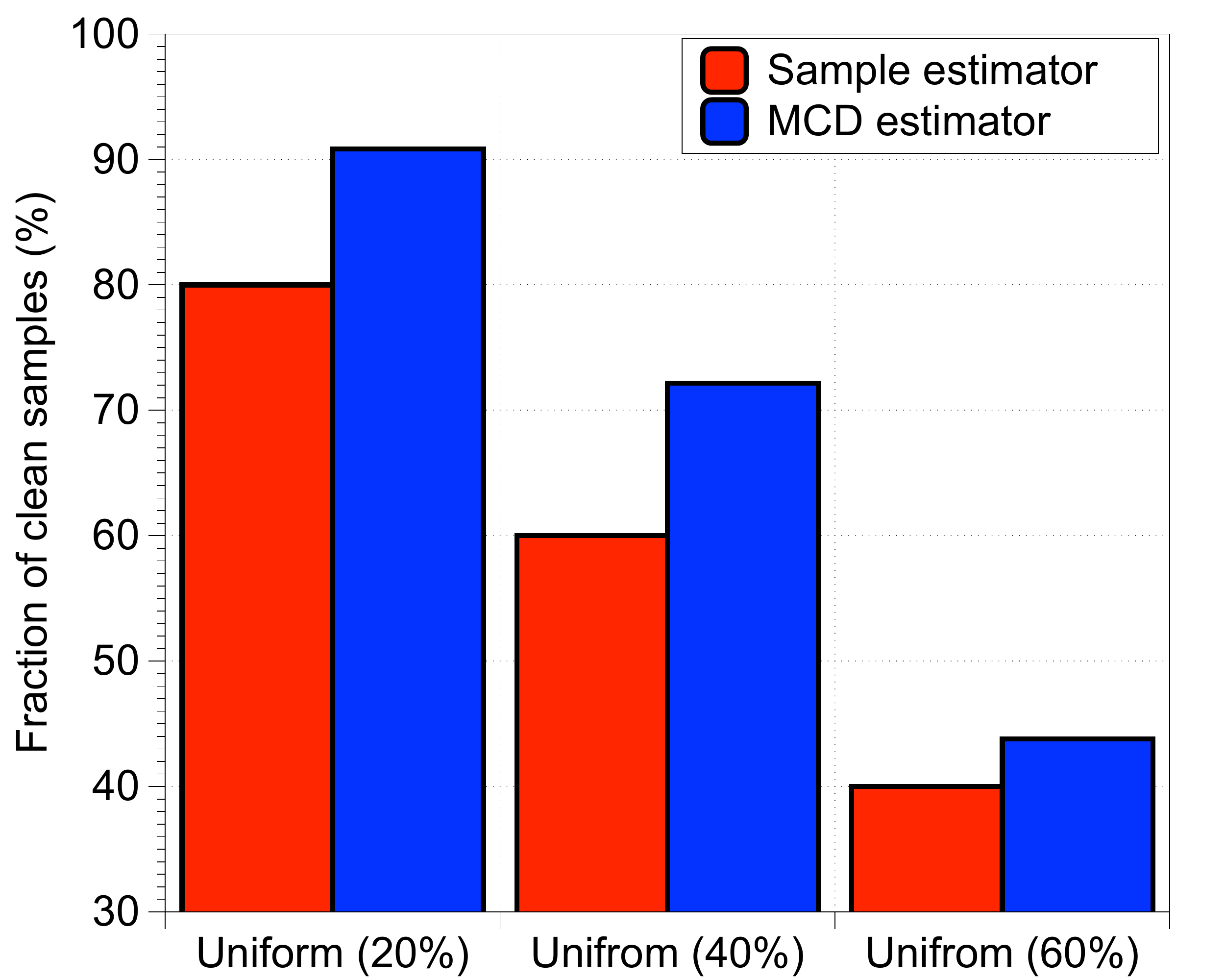} \label{fig:adv_acc_noise}}
\,
\subfigure[Layer-wise accuracy]
{
\includegraphics[width=0.29\textwidth]{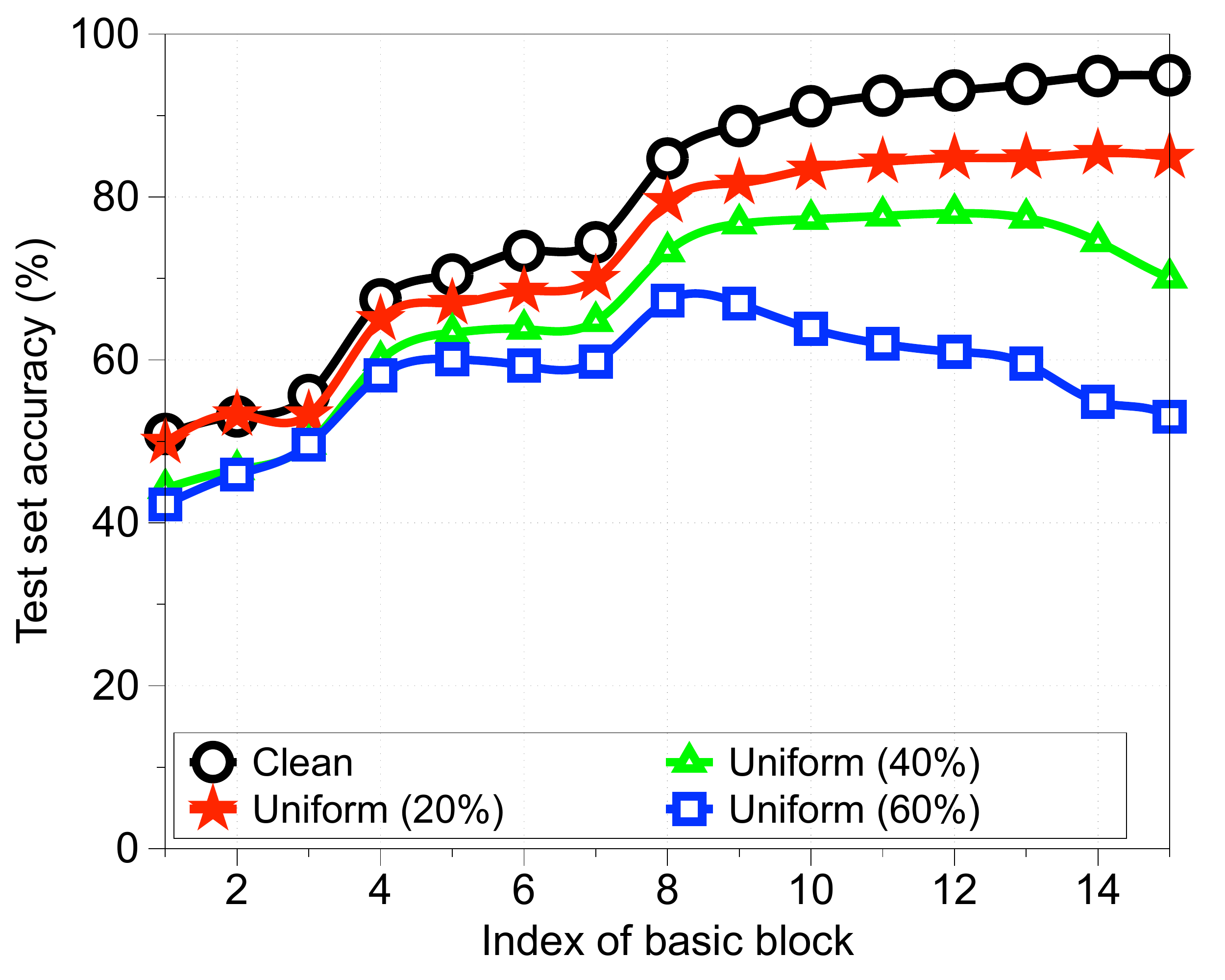} \label{fig:margin}}
\vspace{-0.1in}
\caption{
Experimental results under ResNet-34 model and CIFAR-10 dataset.
(a) Test accuracy of generative classifiers from penultimate features under various assumptions: identity covariance and tied covariance (LDA).
(b) The number of clean samples among selected samples by the MCD estimator.
(c) Test accuracy of generative classifiers computed at different basic blocks.}
\label{fig:analysis}
\end{figure*}

\subsection{Approximation Algorithm for MCD}

Even though the MCD estimator has several advantages, 
the optimization (\ref{eq:mcdopt}) is computationally intractable (i.e., NP-hard) to solve \citep{bernholt2006robust}.
To handle this issue, we aim to compute its approximate solution, by following the idea of \citet{hubert2004fast}.
We design two step scheme as follows: (a) obtain the mean and covariance, i.e., ${\widehat \mu}_c, \mathbf{\widehat \Sigma}_c$, using Algorithm~\ref{alg:MCD_single} for each class $c$, and (b) compute the tied covariance by $\mathbf{\widehat \Sigma} = \frac{\sum_c K_c \mathbf{\widehat \Sigma}_c}{\sum_c K_c}$.
In other words, 
we apply the MCD estimator for each class, and combine the individual covariances into a single one due to the tied covariance assumption of LDA.
Even though finding the optimal solution of the MCD estimator under a single Gaussian distribution is still intractable,
Algorithm~\ref{alg:MCD_single} can produce a local optimal solution since 
it monotonically decreases the determinant under any random initial subset \citep{rousseeuw1999fast}.
We 
choose $I_{\max} = 2$ in our experiments since additional iterations would not improve the results significantly.

\begin{table*}[t]
\small
\centering
\begin{tabular}{cc|c|ccccc}
\toprule
Model & Inference method & Ensemble  & Clean & Uniform (20\%) & Uniform (40\%) & Uniform (60\%) \\ \hline
\multirow{5}{*}{\begin{tabular}[c]{@{}c@{}} DenseNet  \end{tabular}}
& Softmax & - 
& 94.11 & 81.01 & 72.34 & 55.42 \\  \cline{2-7} 
&\multirow{2}{*}{\begin{tabular}[c]{@{}c@{}} Generative + sample \end{tabular}}  
&-
&94.18  & 85.12  & 76.75 & 60.14 \\
&&  \checkmark
& 93.97 & 87.40 &81.27  &  69.81 \\ \cline{2-7} 
&\multirow{2}{*}{\begin{tabular}[c]{@{}c@{}} Generative + MCD (ours)  \end{tabular}}        
& - 
& 94.22 & 86.54 & 80.27 & 67.67 \\
&&  \checkmark   
& 94.18 & {\bf 87.41} & {\bf 81.83} & {\bf 75.45}  \\ 
\midrule
\multirow{5}{*}{\begin{tabular}[c]{@{}c@{}} ResNet  \end{tabular}}
&Softmax &    -             
& 94.76 & 80.88 & 61.98 & 39.96 \\  \cline{2-7} 
&\multirow{2}{*}{\begin{tabular}[c]{@{}c@{}} Generative + sample   \end{tabular}}
& -
& 94.80 & 82.97 & 65.92 & 42.76  \\
&&   \checkmark  
& 94.82 & 83.36 & 68.57 & 46.45 \\ \cline{2-7} 
&\multirow{2}{*}{\begin{tabular}[c]{@{}c@{}} Generative + MCD (ours)  \end{tabular}}        
& - 
& 94.76 & 83.86 & 68.03 & 44.87 \\
&&  \checkmark
& 94.68 & {\bf 84.62} & {\bf 75.28} & {\bf 54.57} \\ 
\bottomrule
\end{tabular}
\caption{Effects of an ensemble method. We use the CIFAR-10 dataset with various uniform noise fractions. All values are percentages and the best results are highlighted in bold if the gain is bigger than 1\% compared to softmax classifier.}
\label{tbl:contributions}
\end{table*}

\subsection{Ensemble of Generative Classifiers}
To further improve the performance of our method,
we consider the ensemble of generative classifiers not only from the penultimate features but also from other low-level features in DNNs.
Formally, given training data, we extract $\ell$-th hidden features of DNNs, denoted by $f_{\ell}(\mathbf{x})\in \mathbb{R}^{d_\ell}$, 
and compute the corresponding parameters of a generative classifier (i.e., ${\widehat \mu}_{\ell,c}$ and $\mathbf{\widehat \Sigma}_\ell$) using the (approximated version of) MCD estimator.
Then, the final posterior distribution is obtained by the weighted sum of all posterior distributions of generative classifiers:
$ \sum \limits_{\ell} \alpha_{\ell} P \left(y=c|f_\ell (\mathbf{x})\right), $
where $\alpha_{\ell}$ is an ensemble weight at $\ell$-th layer.
In our experiments,
we choose the weight of each layer by optimizing negative log-likelihood (NLL) loss over the validation set. 
One can expect that this natural scheme can bring an extra gain in improving the performance due to ensemble effects.

\begin{table*}[t]
\small
\centering
\begin{tabular}{c|ccc|ccc}
\toprule
\multirow{3}{*}{Noise type (\%)} & \multicolumn{3}{c|}{ResNet} & \multicolumn{3}{c}{DenseNet}  \\  \cline{2-7} 
                  & CIFAR-10  & CIFAR-100  & SVHN & CIFAR-10 & CIFAR-100 & SVHN \\ 
                  & \multicolumn{3}{c|}{Softmax / RoG}  &  \multicolumn{3}{c}{Softmax / RoG} \\\hline
Clean             
& 94.76 / 94.68 & 76.81 / 76.97 & 95.96 /  96.09  
& 94.11 / 94.18 & {\bf 75.69} / 72.67 & 96.59 / 96.18 \\ \hline
 {Uniform (20\%)}
& 80.88 / {\bf 84.62} & 64.43 / {\bf 68.29} & 83.52 / {\bf 91.67}    
& 81.01 / {\bf 87.41} & 61.72 / {\bf 64.29} & 86.92 / {\bf 89.50} \\
 {Uniform (40\%)}
& 61.98 / {\bf 75.28} & 48.62 / {\bf 60.76} & 72.89 / {\bf 87.16}   
& 72.34 / {\bf 81.83} & 50.89 / {\bf 55.68} & 81.91 / {\bf 85.71} \\
 {Uniform (60\%)}
& 39.96 / {\bf 54.57} & 27.57 / {\bf 48.42} & 61.23 / {\bf 80.52}  
& 55.42 / {\bf 75.45} & 38.33 / {\bf 44.12} & 71.18 / {\bf 77.67} \\\hline
 {Flip (20\%)}
&  {79.65 / {\bf 88.73}} 
&  {65.14 / {\bf 73.37}} 
&  {85.49 / {\bf 93.00}}     
&  {79.18 / {\bf 91.23}}
&  {65.37 / {\bf 69.03}}
&  {{95.04} / 94.86} \\
 {Flip (40\%)}
&  {58.13 / {\bf 61.56}} 
&  {46.61 / {\bf 66.71}} 
&  {65.88 / {\bf 87.96}}
&  {56.29 / {\bf 86.42}}
&  {46.04 / {\bf 69.38}} 
&  {88.83 / {\bf 93.57}} \\\bottomrule
\end{tabular}
\caption{Test accuracy (\%) of different models trained on various datasets.
We use the ensemble version of RoG, and the best results are highlighted in bold if the gain is bigger than 1\%.}
\label{tbl:main_on_corrupted_label}
\end{table*}

\section{Experiments} \label{sec:experiments}

In this section, we demonstrate the effectiveness of the proposed method using deep neural networks on various vision and natural language processing tasks.
We provide the more detailed experimental setups in the supplementary material.
Code is available at \href{https://github.com/pokaxpoka/RoGNoisyLabel}{\texttt{github.com/pokaxpoka/RoGNoisyLabel}}.

\subsection{Experimental Setup}

For evaluation, we apply the proposed method to deep neural networks including DenseNet \citep{huang2017densely} and ResNet \citep{he2016deep} for the classification tasks on CIFAR \citep{krizhevsky2009learning}, SVHN \citep{netzer2011reading}, Twitter Part of Speech \citep{gimpel2010part}, and Reuters \citep{lewis2004rcv1} datasets with noisy labels.
Following the setups of \citep{ma2018dimensionality,han2018co},
we first consider two types of random noisy labels: corrupting a label to other class uniformly at random (uniform) and corrupting a label only to a specific class (flip). Our method is also evaluated on semantic noisy labels from a machine classifier and open-set noisy labels \citep{wang2018iterative}.


For ensembles of generative classifiers,
we induce the generative classifiers from basic blocks of the last dense (or residual) block of DenseNet (or ResNet), where ensemble weights of each layer are tuned on an additional validation set, 
which consists of 1000 images with noisy labels.
Here, when learning the weights, we use only 500 samples out of 1000, chosen by the MCD estimator to remove the outliers (see the supplementary material for more details).
The size of feature maps on each convolutional layers is reduced by average pooling for computational efficiency: $\mathcal{F} \times \mathcal{H} \times \mathcal{W} \rightarrow \mathcal{F} \times 1$, where $\mathcal{F}$ is the number of channels and $\mathcal{H} \times \mathcal{W}$ is the spatial dimension.

\begin{table*}[t]  
\centering
\resizebox{1.7\columnwidth}{!}{
\begin{tabular}{cc|cccccc}
\toprule
\multirow{2}{*}{Dataset} &
\multirow{2}{*}{\begin{tabular}[c]{@{}c@{}}  Training \\ method \end{tabular}}
& Clean & Uniform (20\%)  
& Uniform (40\%) 
& Uniform (60\%) \\ \cline{3-6} 
\multicolumn{2}{l|}{}                                 
& \multicolumn{4}{c}{Softmax / RoG} \\ \hline
\multirow{6}{*}{CIFAR-10} 
& Cross-entropy
& 94.34 / 94.20
& 81.95 / {\bf 84.63}
& 63.84 / {\bf 74.72} 
& 62.45 / {\bf 67.47} \\
& Bootstrap (hard)
& 94.56 / 94.52
& 82.90 / {\bf 86.27}
& 75.97 / {\bf 80.72} 
& 72.91 / {\bf 75.41} \\
& Bootstrap (soft)
& 94.46 / 94.28
& 80.29 / {\bf 84.82}
& 65.22 / {\bf 74.22} 
& 58.55 / {\bf 66.68} \\
& Forward
&  94.53 / 94.52
&  85.80 / {\bf 86.84}
&  77.95 / {\bf 79.87}
&  72.56 / {\bf 74.75} \\
& Backward
&  94.39 /  94.44
&  77.44 / {\bf 79.16}
&  62.83 / {\bf 68.29}
&  56.64 / {\bf 66.44} \\
& D2L
& 94.55 / 94.29
& 88.89 / 89.00
& 86.68 / 87.00
& 76.83 / {\bf 77.92} \\ \hline
\multirow{6}{*}{CIFAR-100} 
& Cross-entropy
& 76.31 / 75.40
& 61.11 / {\bf 64.82}
& 45.08 / {\bf 55.90}
& 34.97 / {\bf 41.25} \\
& Bootstrap (hard)
& 75.65 / 75.49
& 61.61 / {\bf 64.81}
& 51.27 / {\bf 57.22}
& 39.04 / {\bf 43.69}\\
& Bootstrap (soft)
& 76.40 / 76.02
& 60.28  / {\bf 64.04}
& 47.66 / {\bf 56.51}
& 34.68 / {\bf 42.47}\\
& Forward
& 75.84 / 75.93
& 63.73 /  {\bf 66.02}
& 53.03 /  {\bf 57.69}
& 41.28 /  {\bf 45.28} \\
& Backward
& 76.75 / 76.28
& 56.24 / {\bf 62.13}
& 37.70 / {\bf 50.23}
& 23.55 / {\bf 37.18}\\
& D2L
& 76.13 / 75.93
& 71.90 / 72.09
& 63.61 / {\bf 64.85}
& 9.51 / {\bf 40.57} \\ \hline
\multirow{6}{*}{SVHN} 
& Cross-entropy
&  96.38 / 96.41
&  83.45 / {\bf 91.14}
&  60.86 / {\bf 80.36}
&  38.29 / {\bf 54.99}\\
& Bootstrap (hard)
& 96.40 / 96.12
& 83.43 / {\bf 91.98}
& 74.25 / {\bf 86.83}
& 66.51 / {\bf 77.14}\\
& Bootstrap (soft)
& 96.51 / 96.10
& 86.43 / {\bf 90.84}
& 58.22 / {\bf 79.90}
& 44.52 / {\bf 62.52} \\
& Forward
& 96.36 / 96.00
& 88.21 / {\bf 91.99}
& 80.35 / {\bf 86.49}
& 82.16 / {\bf 84.99} \\
& Backward
& 96.43 / 96.09
& 87.00 /  87.11
& 72.02 /  {\bf 73.32}
& 50.54 /  {\bf 64.01} \\
&  D2L
&  96.49 / 96.37
&  92.31 / {\bf 93.58}
&  94.46 / 94.68
&  92.87 / 93.25 \\
\bottomrule
\end{tabular}}
\caption{
Test accuracy (\%) of ResNet trained on various training methods which utilize a single classifier. We use the ensemble version of RoG, and the best results are highlighted in bold if the gain is bigger than 1\%.}
\label{tbl:comparison_with_SOTA}
\end{table*}

\begin{table*}[t]
\centering
\begin{tabular}{c|c|ccccc}
\toprule
Dataset                   
& Noise type (\%) & Cross-entropy &  {Decoupling} &  {MentorNet} &  {Co-teaching} &
\begin{tabular}[c]{@{}c@{}}   {Co-teaching} \\+   {RoG}\end{tabular}\\ \hline
\multirow{3}{*}{CIFAR-10}  
& Flip (45\%)       
& 49.50 &  48.80 &  58.14 & 71.17 &  71.26 \\
& Uniform (50\%)
& 48.87 & 51.49 &  71.10 &  74.12 &  {\bf 76.67} \\
& Uniform (20\%)   
& 76.25 & 80.44 &  80.76 &  82.13 & {\bf 84.32} \\ \hline
\multirow{3}{*}{CIFAR-100} 
& Flip (45\%)        
& 31.99 &  26.05 &  31.60 & 33.34 & {\bf 43.18} \\
& Uniform (50\%)
& 25.21 &  25.80 & 39.00 & 41.49 & {\bf 45.42} \\
& Uniform (20\%)   
& 47.55 & 44.52  & 52.13 & 54.27 & {\bf 58.16} \\ \bottomrule
\end{tabular}
\caption{Test accuracy (\%) of 9-layer CNNs trained on various training methods which utilize an ensemble of classifiers or meta-learning model. We use the ensemble version of RoG and best results are highlighted in bold if the gain is bigger than 1\%.}
\label{tbl:main_on_co-teaching}
\end{table*}

\begin{table*}[t] 
\centering
\resizebox{1.7\columnwidth}{!}{
\begin{tabular}{cc|ccccc}
\toprule
\multirow{2}{*}{Dataset} &
\multirow{2}{*}{\begin{tabular}[c]{@{}c@{}}Training \\method \end{tabular}}
& \multicolumn{4}{c}{Softmax / RoG}\\ \cline{3-6} 
\multicolumn{2}{l|}{}   
& Clean & Uniform (20\%) & Uniform (40\%) & Uniform (60\%) 
\\ \hline
\multirow{3}{*}{Twitter} 
& Cross-entropy
& {\bf 87.47} / 85.28
& 79.13 / {\bf 81.66}
& 66.74 / {\bf 79.37}
& 50.83 / {\bf 73.65} \\
& Forward (gold) 
& 78.07 / {\bf 83.59}
& 72.97 / {\bf 81.60}
& 64.55 / {\bf 78.24}
& 51.59 / {\bf 72.33} \\
& GLC 
& 83.47 / {\bf 84.68}
& 66.09 / {\bf 81.66}
& 59.72 / {\bf 79.00}
& 53.14 / {\bf 72.93} \\ \midrule
\multirow{3}{*}{Reuters} 
& Cross-entropy
& {\bf 95.88} / 94.77
& 87.74 / {\bf 92.83}
& 76.54 / {\bf 82.20}
& 57.49 / {\bf 64.98} \\
& Forward (gold) 
& 94.57 / 94.75
& 88.44 / {\bf 93.24}
& 77.85 / {\bf 82.56}
& 61.01 / {\bf 66.56} \\
& GLC 
& {\bf 95.97} / 94.91
& 81.45 / {\bf 92.75}
& 73.40 / {\bf 83.82}
& 59.21 / {\bf 67.91} \\
\bottomrule
\end{tabular}}
\caption{Test accuracy (\%) of 2-layer FCNs trained on NLP datasets with uniform noise. We use the ensemble version of RoG, and the best results highlighted in hold if the gain is bigger than 1\%.}
\label{tbl:comparison_with_SOTA_nlp}
\vspace*{-0.1in}
\end{table*}

\subsection{Ablation Study}
We first evaluate the performance of generative classifiers with various assumptions: identity covariance (Euclidean) and tied covariance (LDA).
In the case of identity covariance, we also apply a robust estimator called the least trimmed square (LTS) estimator \citep{rousseeuw1984least} which finds a $K$-subset with smallest error and computes the sample mean from it, i.e., $\min_{\widehat \mu} \sum_{i=1}^{K} (\| \mathbf{x}_i - \widehat \mu \|_2^2)$.
Figure~\ref{fig:l_acc_noise} reports the test set accuracy of the softmax and generative classifiers on features extracted from the penultimate layer using ResNet-34 trained on the CIFAR-10 dataset with the uniform noise type.
First, one can observe that the generative classifiers with LDA assumption (blue and purple bars) generalize better than the softmax (red bar) and generative classifiers with identity covariance (orange and green bars) well from noisy labels.
Here, we remark that they still provide a comparable classification accuracy of softmax classifier when the model is trained on clean dataset (i.e., noise = 0\%).
On top of that, by utilizing the MCD estimator, the classification accuracy (blue bar) is further improved compared to that employs only the naive sample estimator (purple bar).
This is because the MCD estimator indeed selects the training samples with clean labels as shown in Figure~\ref{fig:adv_acc_noise}.
The above results justify the proposed generative classifier, in comparison with other alternatives.


Next, to confirm that the ensemble approach is indeed effective,
we measure a classification accuracy of generative classifier from different basic blocks of ResNet-34.
First, we found that the performances of the generative classifiers from low-level features are more stable, while the accuracy of generative classifier from penultimate layer significantly decreases as the noisy fraction increases as shown in Figure~\ref{fig:margin}.
We expect that this is because the dimension (i.e., number of channels) of low-level features is usually smaller than that of high-level features.
Since the breakdown point of MCD is inversely proportional to the feature dimension, the generative classifiers from low-level features can be more robust.
This also coincides with the prior observation in the literature \citep{morcos2018insights} that DNNs tend to have similar hidden features at early layers, regardless of whether they train clean or noisy labels.
Since the generative classifiers from low-level features are more stable, the ensemble method significantly improves the classification accuracy as shown in Table~\ref{tbl:contributions}.
Finally, Table~\ref{tbl:main_on_corrupted_label} reports the classification accuracy for all networks and datasets, where the proposed method significantly outperforms the softmax classifier for all tested cases.

\subsection{Compatibility and Comparison with the State-of-Art Training Methods}

We compare the performance of the standard softmax classifier and RoG when they are combined with other various training methods for noisy environments, where more detailed explanations about training methods are given in the supplementary material. 
First, we consider the following methods that require to train only a single network: Hard/soft bootstrapping \citep{reed2014training}, forward/backward \citep{patrini2017making}, and D2L \citep{ma2018dimensionality}.
Following the same experimental setup in \citet{ma2018dimensionality}\footnote{The code is available at \url{https://github.com/xingjunm/dimensionality-driven-learning}.},
we use ResNet-44 and only consider the uniform noises of various levels.
Table~\ref{tbl:comparison_with_SOTA} shows the classification accuracy of softmax classifier and the ensemble version of RoG. Note that RoG always improves the classification accuracy compared to the softmax classifier, where the gains due to ours are more significant than those due to other special training methods.

We also consider the following methods that require to train multiple networks, i.e., an ensemble of classifiers or a meta-learning model: Decoupling \citep{malach2017decoupling}, MentorNet \citep{jiang2017mentornet} and Co-teaching \citep{han2018co}.
Following the same experimental setup of \citet{han2018co}\footnote{We used a reference implementation: \url{https://github.com/bhanML/Co-teaching}.},
we use a 9-layer convolutional neural network (CNN), and consider the CIFAR-10 and CIFAR-100 datasets with uniform and flip noise.
In this setup, we only apply RoG to a model pre-trained by Co-teaching since it outperforms other training methods.
As shown in Table~\ref{tbl:main_on_co-teaching}, RoG with Co-teaching method achieves the best performance in all tested cases.

\begin{table*}[t]  
\small
\centering
\begin{tabular}{c|ccc|ccc}
\toprule
\multirow{3}{*}{\begin{tabular}[c]{@{}c@{}}  Training \\ method \end{tabular}}
& \multicolumn{3}{c|}{Label generator (noisy fraction) on CIFAR-10}
& \multicolumn{3}{c}{Label generator (noisy fraction) on CIFAR-100}\\ 
& DenseNet (32\%)   
& ResNet (38\%)
& VGG (34\%) 
& DenseNet (34\%)   
& ResNet (37\%)
& VGG (37\%) \\  \cline{2-7} 
& \multicolumn{3}{c|}{Softmax / RoG} 
& \multicolumn{3}{c}{Softmax / RoG} \\ \hline
Cross-entropy
& 67.24 / {\bf 68.33}
& 62.26 / {\bf 64.15} 
& 68.77 / {\bf 70.04}
& 50.72 / {\bf 61.14}
& 50.68 / {\bf 53.09}
& 51.08 / {\bf 53.64}
\\
Bootstrap (hard)
& 67.31 / {\bf 68.40}
& 62.22 / {\bf 63.98} 
& 69.11 / {\bf 70.09} 
& 51.31 / {\bf 53.66}
& 50.62 / {\bf 52.62}
& 50.91 / {\bf 53.46} \\
Bootstrap (soft)
& 67.17 / {\bf 68.38}
& 62.15 / {\bf 64.03} 
& 69.28 / {\bf 70.11}
& 50.57 / {\bf 54.71}
& 50.68 / {\bf 53.30}
& 51.41 / {\bf 53.76}\\
Forward
&  67.46 / {\bf 68.20}
&  61.96 / {\bf 64.24} 
&  68.95 / {\bf 70.09} 
& 50.59 / {\bf 53.91}
& 51.04  / {\bf 53.36}
& 51.05 / {\bf 53.63} \\
Backward
& 67.31 / {\bf 68.66}
& 62.40 / {\bf 63.45} 
& 69.04 / {\bf 70.18}
& 50.54 / {\bf 54.01}
& 50.30 / {\bf 53.03}
& 51.15 / {\bf 53.50} \\
D2L
& 66.91 / {\bf 68.57}
& 59.10 / {\bf 60.25} 
& 57.97 / {\bf 59.94}
& 5.00 / {\bf 31.67}
& 23.71 / {\bf 39.92}
& 40.97 / {\bf 45.42}\\
\bottomrule
\end{tabular}
\caption{Test accuracy (\%) of DenseNet on the CIFAR-10 and CIFAR-100 datasets with semantic noisy labels. We use the ensemble version of RoG, and the best results are highlighted in bold if the gain is bigger than 1\%.}
\label{tbl:comparison_with_SOTA_temp_dense}
\vspace*{-0.1in}
\end{table*}

We further apply 
our inference method to non-convolutional neural networks on natural language processing (NLP) tasks: 
the text categorization on Reuters \citep{lewis2004rcv1}, and part-of-speech (POS) tagging on Twitter POS \citep{gimpel2010part}.
By following the same experimental setup of \citet{hendrycks2018using}\footnote{The code is available at \url{https://github.com/mmazeika/glc}.},
we train 2-layer fully connected networks (FCNs) using forward (gold)\footnote{We use an augmented version of forward \citep{patrini2017making} which estimates a corruption matrix using the trusted data.},
and GLC \citep{hendrycks2018using} methods.
Note that they are designed to train a single network using a set of trusted data with golden clean labels {(1\% of training samples in our experiments).}
Hence, the setting is slightly different from what we considered so far,
but we run RoG (without utilizing 1\% knowledge of ground truth) to compare. 
Table~\ref{tbl:comparison_with_SOTA_nlp} shows that, even with this unfair disadvantage, RoG can improve the performance over the baselines for these NLP datasets with noisy labels.

\begin{table}[t]
\centering
\small
\begin{tabular}{c|cc}
\toprule
Open-set data & Softmax & RoG\\ \midrule
 CIFAR-100 
& 79.01 & {\bf 83.37} \\
 ImageNet  
& 86.88 & 87.05 \\
 CIFAR-100 + ImageNet  
& 81.58 & {\bf 84.35} \\ \bottomrule
\end{tabular}
\caption{
Test accuracy (\%) of DenseNet on the CIFAR-10 dataset with open-set noisy labels. We use the ensemble of RoG. The best results are highlighted in bold if the gain is bigger than 1\%.}
\label{tbl:openset_noise}
\vspace*{-0.15in}
\end{table}

\subsection{Semantic and Open-Set Noisy Labels} \label{sec:semantic}

\begin{figure}[t]
\centering
\subfigure[Confusion graph on CIFAR-10]
{
\includegraphics[width=0.4\textwidth]{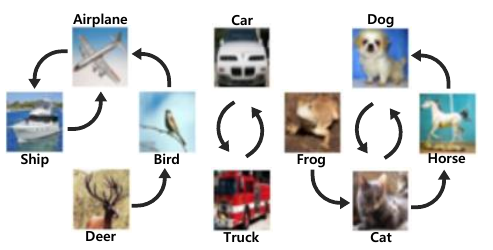} \label{fig:confusion}} 
\,
\subfigure[Open-set noise example]
{
\includegraphics[width=0.38\textwidth]{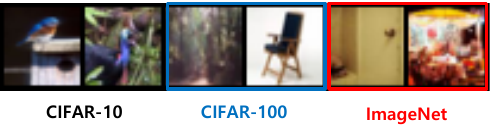} \label{fig:openset}}
\caption{(a) Confusion graph from ResNet-34 trained on the CIFAR-10 dataset. (b) Examples of training samples with open-set noise for ``bird" in the CIFAR-10 dataset.}\label{fig:openset_and_confusion}
\end{figure}

In this section, our method is evaluated under more realistic noisy labels. First, in order to generate more semantically meaningful noisy labels, 
we train DenseNet-100, ResNet-34 and VGG-13 \citep{simonyan2014very} using 5\% and 20\% of CIFAR-10 and CIFAR-100 training samples with clean labels, respectively. 
Then, we produce the labels of remaining training samples based on their predictions, 
and train another DenseNet-100 with the pseudo-labeled samples.
{Figure~\ref{fig:confusion} shows a confusion graph for pseudo-labels, obtained from ResNet-34 trained on the 5\% of CIFAR-10: each node corresponds to a class, and an edge from the node represents its most confusing class.
Note that the weak classification system produces semantically  noisy labels; 
e.g., ``Cat" is confused with ``Dog", but not with ``Car''. 
We remark that DenseNet and VGG also produce similar confusion graphs.
Table~\ref{tbl:comparison_with_SOTA_temp_dense} shows RoG consistently improves the performance, while the gains due to other special training methods are not very significant.}

Our final benchmark is the open-set noisy scenario \citep{wang2018iterative}.
In this case, some training images are often from the open world
and not relevant to the targeted classification task at all, i.e., out-of-distribution.
However, they are still labeled to certain classes.
For example, as shown in Figure~\ref{fig:openset},
noisy samples like ``chair" from CIFAR-100 and ``door" from (downsampled) ImageNet \citep{chrabaszcz2017downsampled} can be labeled as ``bird" to train, even though their true labels are not contained within the set of training classes in the CIFAR-10 dataset.
In our experiments,
open-set noisy datasets are built by replacing some training samples in CIFAR-10 by out-of-distribution samples, while keeping the labels and the number of images per class unchanged.
We train DenseNet-100 on CIFAR-10 with 60\% open-set noise. 
As shown in Table~\ref{tbl:openset_noise}, our method achieves comparable or significantly better test accuracy than the softmax classifier.

\section{Conclusion} \label{sec:conclusion}

We propose a new inference method for handling noisy labels.
Our main idea is inducing the generative classifier on top of fixed features from the pre-trained model.
Such ``deep generative classifiers'' have been largely dismissed for fully-supervised classification settings as they are often substantially outperformed by discriminative deep classifiers (e.g., softmax classifiers). In contrast to this common belief, we show that it is possible to formulate a simple generative classifier that is significantly more robust to labeling noise without much sacrifice of the discriminative performance for clean labeling setting. 
We expect that our work would bring a refreshing perspective for other related tasks, e.g., memorization \citep{zhang2016understanding}, adversarial attacks \citep{szegedy2013intriguing}, and semi-supervised learning \citep{oliver2018realistic}.

\section*{Acknowledgements}
This work was supported by Institute for Information \& communications Technology Planning \& Evaluation(IITP) grant funded by the Korea government(MSIT) (No.2017-0-01779, A machine learning and statistical inference framework for explainable artificial intelligence), Sloan Research Fellowship, and Kwanjeong Educational Foundation Scholarship. This work is partially supported by DARPA under Grant 00009970, Wechat. We also thank Dawn Song, Dan
Hendrycks, Sungsoo Ahn and Insu Han for helpful discussions.

\bibliography{icml2019_reference}

\begin{thebibliography}{45}
\providecommand{\natexlab}[1]{#1}
\providecommand{\url}[1]{\texttt{#1}}
\expandafter\ifx\csname urlstyle\endcsname\relax
  \providecommand{\doi}[1]{doi: #1}\else
  \providecommand{\doi}{doi: \begingroup \urlstyle{rm}\Url}\fi

\bibitem[Amodei et~al.(2016)Amodei, Ananthanarayanan, Anubhai, Bai, Battenberg,
  Case, Casper, Catanzaro, Cheng, Chen, et~al.]{amodei2016deep}
Amodei, D., Ananthanarayanan, S., Anubhai, R., Bai, J., Battenberg, E., Case,
  C., Casper, J., Catanzaro, B., Cheng, Q., Chen, G., et~al.
\newblock Deep speech 2: End-to-end speech recognition in english and mandarin.
\newblock In \emph{ICML}, 2016.

\bibitem[Arpit et~al.(2017)Arpit, Jastrzebski, Ballas, Krueger, Bengio, Kanwal,
  Maharaj, Fischer, Courville, Bengio, et~al.]{arpit2017closer}
Arpit, D., Jastrzebski, S., Ballas, N., Krueger, D., Bengio, E., Kanwal, M.~S.,
  Maharaj, T., Fischer, A., Courville, A., Bengio, Y., et~al.
\newblock A closer look at memorization in deep networks.
\newblock In \emph{ICML}, 2017.

\bibitem[Bernholt(2006)]{bernholt2006robust}
Bernholt, T.
\newblock Robust estimators are hard to compute.
\newblock Technical report, Technical Report/Universit{\"a}t Dortmund, SFB 475
  Komplexit{\"a}tsreduktion in Multivariaten Datenstrukturen, 2006.

\bibitem[Chrabaszcz et~al.(2017)Chrabaszcz, Loshchilov, and
  Hutter]{chrabaszcz2017downsampled}
Chrabaszcz, P., Loshchilov, I., and Hutter, F.
\newblock A downsampled variant of imagenet as an alternative to the cifar
  datasets.
\newblock \emph{arXiv preprint arXiv:1707.08819}, 2017.

\bibitem[Deng et~al.(2009)Deng, Dong, Socher, Li, Li, and
  Fei-Fei]{deng2009imagenet}
Deng, J., Dong, W., Socher, R., Li, L.-J., Li, K., and Fei-Fei, L.
\newblock Imagenet: A large-scale hierarchical image database.
\newblock In \emph{CVPR}, 2009.

\bibitem[Durrant \& Kab{\'a}n(2010)Durrant and
  Kab{\'a}n]{durrant2010compressed}
Durrant, R.~J. and Kab{\'a}n, A.
\newblock Compressed fisher linear discriminant analysis: Classification of
  randomly projected data.
\newblock In \emph{ACM SIGKDD}, 2010.

\bibitem[Fisher(1936)]{fisher1936use}
Fisher, R.~A.
\newblock The use of multiple measurements in taxonomic problems.
\newblock \emph{Annals of eugenics}, 1936.

\bibitem[Garcia-Escudero \& Gordaliza(1999)Garcia-Escudero and
  Gordaliza]{garcia1999robustness}
Garcia-Escudero, L.~A. and Gordaliza, A.
\newblock Robustness properties of k means and trimmed k means.
\newblock \emph{Journal of the American Statistical Association}, 1999.

\bibitem[Gimpel et~al.(2011)Gimpel, Schneider, O'Connor, Das, Mills,
  Eisenstein, Heilman, Yogatama, Flanigan, and Smith]{gimpel2010part}
Gimpel, K., Schneider, N., O'Connor, B., Das, D., Mills, D., Eisenstein, J.,
  Heilman, M., Yogatama, D., Flanigan, J., and Smith, N.~A.
\newblock Part-of-speech tagging for twitter: Annotation, features, and
  experiments.
\newblock In \emph{ACL}, 2011.

\bibitem[Girshick(2015)]{girshick2015fast}
Girshick, R.
\newblock Fast r-cnn.
\newblock In \emph{ICCV}, 2015.

\bibitem[Goldberger \& Ben-Reuven(2017)Goldberger and
  Ben-Reuven]{goldberger2016training}
Goldberger, J. and Ben-Reuven, E.
\newblock Training deep neural-networks using a noise adaptation layer.
\newblock In \emph{ICLR}, 2017.

\bibitem[Hampel(1971)]{hampel1971general}
Hampel, F.~R.
\newblock A general qualitative definition of robustness.
\newblock \emph{The Annals of Mathematical Statistics}, 1971.

\bibitem[Han et~al.(2018{\natexlab{a}})Han, Niu, Yao, Yu, Xu, Tsang, and
  Sugiyama]{han2018pumpout}
Han, B., Niu, G., Yao, J., Yu, X., Xu, M., Tsang, I., and Sugiyama, M.
\newblock Pumpout: A meta approach for robustly training deep neural networks
  with noisy labels.
\newblock \emph{arXiv preprint arXiv:1809.11008}, 2018{\natexlab{a}}.

\bibitem[Han et~al.(2018{\natexlab{b}})Han, Yao, Yu, Niu, Xu, Hu, Tsang, and
  Sugiyama]{han2018co}
Han, B., Yao, Q., Yu, X., Niu, G., Xu, M., Hu, W., Tsang, I., and Sugiyama, M.
\newblock Co-teaching: robust training deep neural networks with extremely
  noisy labels.
\newblock In \emph{NeurIPS}, 2018{\natexlab{b}}.

\bibitem[He et~al.(2016)He, Zhang, Ren, and Sun]{he2016deep}
He, K., Zhang, X., Ren, S., and Sun, J.
\newblock Deep residual learning for image recognition.
\newblock In \emph{CVPR}, 2016.

\bibitem[Hendrycks et~al.(2018)Hendrycks, Mazeika, Wilson, and
  Gimpel]{hendrycks2018using}
Hendrycks, D., Mazeika, M., Wilson, D., and Gimpel, K.
\newblock Using trusted data to train deep networks on labels corrupted by
  severe noise.
\newblock In \emph{NeurIPS}, 2018.

\bibitem[Hermansky et~al.(2000)Hermansky, Ellis, and
  Sharma]{hermansky2000tandem}
Hermansky, H., Ellis, D.~P., and Sharma, S.
\newblock Tandem connectionist feature extraction for conventional hmm systems.
\newblock In \emph{icassp}, 2000.

\bibitem[Huang \& Liu(2017)Huang and Liu]{huang2017densely}
Huang, G. and Liu, Z.
\newblock Densely connected convolutional networks.
\newblock In \emph{CVPR}, 2017.

\bibitem[Hubert \& Van~Driessen(2004)Hubert and Van~Driessen]{hubert2004fast}
Hubert, M. and Van~Driessen, K.
\newblock Fast and robust discriminant analysis.
\newblock \emph{Computational Statistics \& Data Analysis}, 2004.

\bibitem[Jiang et~al.(2018)Jiang, Zhou, Leung, Li, and
  Fei-Fei]{jiang2017mentornet}
Jiang, L., Zhou, Z., Leung, T., Li, L.-J., and Fei-Fei, L.
\newblock Mentornet: Regularizing very deep neural networks on corrupted
  labels.
\newblock In \emph{ICML}, 2018.

\bibitem[Krause et~al.(2016)Krause, Sapp, Howard, Zhou, Toshev, Duerig,
  Philbin, and Fei-Fei]{krause2016unreasonable}
Krause, J., Sapp, B., Howard, A., Zhou, H., Toshev, A., Duerig, T., Philbin,
  J., and Fei-Fei, L.
\newblock The unreasonable effectiveness of noisy data for fine-grained
  recognition.
\newblock In \emph{ECCV}, 2016.

\bibitem[Krizhevsky \& Hinton(2009)Krizhevsky and
  Hinton]{krizhevsky2009learning}
Krizhevsky, A. and Hinton, G.
\newblock Learning multiple layers of features from tiny images.
\newblock \emph{Technical report, University of Toronto}, 2009.

\bibitem[Kuznetsova et~al.(2018)Kuznetsova, Rom, Alldrin, Uijlings, Krasin,
  Pont-Tuset, Kamali, Popov, Malloci, Duerig, et~al.]{kuznetsova2018open}
Kuznetsova, A., Rom, H., Alldrin, N., Uijlings, J., Krasin, I., Pont-Tuset, J.,
  Kamali, S., Popov, S., Malloci, M., Duerig, T., et~al.
\newblock The open images dataset v4: Unified image classification, object
  detection, and visual relationship detection at scale.
\newblock \emph{arXiv preprint arXiv:1811.00982}, 2018.

\bibitem[Lasserre et~al.(2006)Lasserre, Bishop, and
  Minka]{lasserre2006principled}
Lasserre, J.~A., Bishop, C.~M., and Minka, T.~P.
\newblock Principled hybrids of generative and discriminative models.
\newblock In \emph{CVPR}, 2006.

\bibitem[Lee et~al.(2018)Lee, Lee, Lee, and Shin]{lee2018simple}
Lee, K., Lee, K., Lee, H., and Shin, J.
\newblock A simple unified framework for detecting out-of-distribution samples
  and adversarial attacks.
\newblock In \emph{NeurIPS}, 2018.

\bibitem[Lewis et~al.(2004)Lewis, Yang, Rose, and Li]{lewis2004rcv1}
Lewis, D.~D., Yang, Y., Rose, T.~G., and Li, F.
\newblock Rcv1: A new benchmark collection for text categorization research.
\newblock \emph{JMLR}, 2004.

\bibitem[Lopuhaa et~al.(1991)Lopuhaa, Rousseeuw, et~al.]{lopuhaa1991breakdown}
Lopuhaa, H.~P., Rousseeuw, P.~J., et~al.
\newblock Breakdown points of affine equivariant estimators of multivariate
  location and covariance matrices.
\newblock \emph{The Annals of Statistics}, 1991.

\bibitem[Ma et~al.(2018)Ma, Wang, Houle, Zhou, Erfani, Xia, Wijewickrema, and
  Bailey]{ma2018dimensionality}
Ma, X., Wang, Y., Houle, M.~E., Zhou, S., Erfani, S.~M., Xia, S.-T.,
  Wijewickrema, S., and Bailey, J.
\newblock Dimensionality-driven learning with noisy labels.
\newblock In \emph{ICML}, 2018.

\bibitem[Maaten \& Hinton(2008)Maaten and Hinton]{maaten2008visualizing}
Maaten, L. v.~d. and Hinton, G.
\newblock Visualizing data using t-sne.
\newblock \emph{Journal of machine learning research}, 2008.

\bibitem[Mahajan et~al.(2018)Mahajan, Girshick, Ramanathan, He, Paluri, Li,
  Bharambe, and van~der Maaten]{mahajan2018exploring}
Mahajan, D., Girshick, R., Ramanathan, V., He, K., Paluri, M., Li, Y.,
  Bharambe, A., and van~der Maaten, L.
\newblock Exploring the limits of weakly supervised pretraining.
\newblock In \emph{ECCV}, 2018.

\bibitem[Malach \& Shalev-Shwartz(2017)Malach and
  Shalev-Shwartz]{malach2017decoupling}
Malach, E. and Shalev-Shwartz, S.
\newblock Decoupling" when to update" from" how to update".
\newblock In \emph{NeurIPS}, 2017.

\bibitem[Morcos et~al.(2018)Morcos, Raghu, and Bengio]{morcos2018insights}
Morcos, A.~S., Raghu, M., and Bengio, S.
\newblock Insights on representational similarity in neural networks with
  canonical correlation.
\newblock In \emph{NeurIPS}, 2018.

\bibitem[Netzer et~al.(2011)Netzer, Wang, Coates, Bissacco, Wu, and
  Ng]{netzer2011reading}
Netzer, Y., Wang, T., Coates, A., Bissacco, A., Wu, B., and Ng, A.~Y.
\newblock Reading digits in natural images with unsupervised feature learning.
\newblock In \emph{NeurIPS workshop}, 2011.

\bibitem[Ng \& Jordan(2002)Ng and Jordan]{ng2002discriminative}
Ng, A.~Y. and Jordan, M.~I.
\newblock On discriminative vs. generative classifiers: A comparison of
  logistic regression and naive bayes.
\newblock In \emph{NeurIPS}, 2002.

\bibitem[Oliver et~al.(2018)Oliver, Odena, Raffel, Cubuk, and
  Goodfellow]{oliver2018realistic}
Oliver, A., Odena, A., Raffel, C., Cubuk, E.~D., and Goodfellow, I.~J.
\newblock Realistic evaluation of deep semi-supervised learning algorithms.
\newblock In \emph{NeurIPS}, 2018.

\bibitem[Patrini et~al.(2017)Patrini, Rozza, Menon, Nock, and
  Qu]{patrini2017making}
Patrini, G., Rozza, A., Menon, A.~K., Nock, R., and Qu, L.
\newblock Making deep neural networks robust to label noise: A loss correction
  approach.
\newblock In \emph{CVPR}, 2017.

\bibitem[Reed et~al.(2014)Reed, Lee, Anguelov, Szegedy, Erhan, and
  Rabinovich]{reed2014training}
Reed, S., Lee, H., Anguelov, D., Szegedy, C., Erhan, D., and Rabinovich, A.
\newblock Training deep neural networks on noisy labels with bootstrapping.
\newblock \emph{arXiv preprint arXiv:1412.6596}, 2014.

\bibitem[Ren et~al.(2018)Ren, Zeng, Yang, and Urtasun]{ren2018learning}
Ren, M., Zeng, W., Yang, B., and Urtasun, R.
\newblock Learning to reweight examples for robust deep learning.
\newblock In \emph{ICML}, 2018.

\bibitem[Rousseeuw(1984)]{rousseeuw1984least}
Rousseeuw, P.~J.
\newblock Least median of squares regression.
\newblock \emph{Journal of the American statistical association}, 1984.

\bibitem[Rousseeuw \& Driessen(1999)Rousseeuw and Driessen]{rousseeuw1999fast}
Rousseeuw, P.~J. and Driessen, K.~V.
\newblock A fast algorithm for the minimum covariance determinant estimator.
\newblock \emph{Technometrics}, 1999.

\bibitem[Simonyan \& Zisserman(2015)Simonyan and Zisserman]{simonyan2014very}
Simonyan, K. and Zisserman, A.
\newblock Very deep convolutional networks for large-scale image recognition.
\newblock In \emph{ICLR}, 2015.

\bibitem[Szegedy et~al.(2014)Szegedy, Zaremba, Sutskever, Bruna, Erhan,
  Goodfellow, and Fergus]{szegedy2013intriguing}
Szegedy, C., Zaremba, W., Sutskever, I., Bruna, J., Erhan, D., Goodfellow, I.,
  and Fergus, R.
\newblock Intriguing properties of neural networks.
\newblock In \emph{ICLR}, 2014.

\bibitem[Wang et~al.(2018)Wang, Liu, Ma, Bailey, Zha, Song, and
  Xia]{wang2018iterative}
Wang, Y., Liu, W., Ma, X., Bailey, J., Zha, H., Song, L., and Xia, S.-T.
\newblock Iterative learning with open-set noisy labels.
\newblock In \emph{CVPR}, 2018.

\bibitem[Zhang et~al.(2017)Zhang, Bengio, Hardt, Recht, and
  Vinyals]{zhang2016understanding}
Zhang, C., Bengio, S., Hardt, M., Recht, B., and Vinyals, O.
\newblock Understanding deep learning requires rethinking generalization.
\newblock In \emph{ICLR}, 2017.

\bibitem[Zhang \& Sabuncu(2018)Zhang and Sabuncu]{zhang2018generalized}
Zhang, Z. and Sabuncu, M.~R.
\newblock Generalized cross entropy loss for training deep neural networks with
  noisy labels.
\newblock In \emph{NeurIPS}, 2018.

\end{thebibliography}
\bibliographystyle{icml2019}

\appendix
\onecolumn
\clearpage
\begin{center}{\bf {\LARGE Supplementary Material:}}
\end{center}

\begin{center}{\bf {\Large Robust Inference via Generative Classifiers for Handling Noisy Labels}}
\end{center}

\section{Preliminaries} \label{appendix:preliminaries}

{\bf Gaussian discriminant analysis}. In this section, we describe the basic concepts of the discriminative and generative classifier \citep{ng2002discriminative}.
Formally, denote the random variable of the input and label as $\mathbf{x}$ and $y  = \{1,\cdots,C\}$, respectively.
For the classification task,
the discriminative classifier directly defines a posterior distribution $P(y|\mathbf{x})$, i.e., learning a direct mapping between input $\mathbf{x}$ and label $y$. A popular model for discriminative classifier is softmax classifier which defines the posterior distribution as follows:
$P\left(y=c|\mathbf{x}\right) = \frac{ \exp \left( \mathbf{w}_c^\top \mathbf{x} + b_c \right) }{\sum_{c^\prime} \exp \left( \mathbf{w}_{c^\prime}^\top \mathbf{x} + b_{c^\prime}\right) },$ 
where $\mathbf{w}_c$ and $b_c$ are weights and bias for a class $c$, respectively. 
In contrast to the discriminative classifier, the generative classifier defines the class conditional distribution $P\left( \mathbf{x}|y\right)$ and class prior $P\left( y \right)$ in order to indirectly define the posterior distribution by specifying the joint distribution $P\left(\mathbf{x}, y\right) = P\left( y \right) P \left( \mathbf{x}| y \right)$.
Gaussian discriminant analysis (GDA) is a popular method to define the generative classifier by assuming that
the class conditional distribution follows the multivariate Gaussian distribution and the class prior follows Bernoulli distribution:
$P\left( \mathbf{x}|y=c\right) = \mathcal{N} \left(\mathbf{x}| \mathbf{\mu}_c, \mathbf{\Sigma}_c \right),
P\left( y=c\right) = \frac{\beta_c}{\sum_{c^\prime} \beta_{c^\prime}},$
where $\mathbf{\mu}_c$ and $\mathbf{\Sigma}_c$ are the mean and covariance of multivariate Gaussian distribution, and $\beta_c$ is the unnormalized prior for class $c$.
This classifier has been studied in various machine learning areas (e.g., semi-supervised learning \citep{lasserre2006principled} and incremental learning \citep{lee2018simple}).

In this paper, we focus on the special case of GDA, also known as the linear discriminant analysis (LDA). In addition to Gaussian assumption, LDA further assumes that all classes share the same covariance matrix, i.e., $\mathbf{\Sigma}_c = \mathbf{\Sigma}$. 
Since the quadratic term is canceled out with this assumption,
the posterior distribution of generative classifier can be represented as follows:
\begin{align*}
& P\left(y=c|\mathbf{x}\right) = \frac{P\left( y = c \right) P\left(\mathbf{x}| y =c \right) }{\sum_{c^\prime}P\left( y= c^\prime \right) P\left(\mathbf{x}| y= c^\prime \right)} \notag = \frac{ \exp \left( \mathbf{\mu}_c^\top \mathbf{\Sigma}^{-1} \mathbf{x} -\frac{1}{2} \mathbf{\mu}_c^\top \mathbf{\Sigma}^{-1} \mathbf{\mu}_c +\log \beta_c \right) }{\sum_{c^\prime} \exp \left( \mathbf{\mu}_{c^\prime}^\top \mathbf{\Sigma}^{-1} \mathbf{x} -\frac{1}{2} \mathbf{\mu}_{c^\prime}^\top \mathbf{\Sigma}^{-1} \mathbf{\mu}_{c^\prime} +\log \beta_{c^\prime} \right)}.
\end{align*}
One can note that the above form of posterior distribution is equivalent to the softmax classifier by considering $\mathbf{\mu}_{c}^\top \mathbf{\Sigma}^{-1}$ and $ -\frac{1}{2} \mathbf{\mu}_c^\top \mathbf{\Sigma}^{-1} \mathbf{\mu}_c +\log \beta_c$ as its weight and bias, respectively. This implies that $\mathbf{x}$ might be fitted in Gaussian distribution during training a softmax classifier.

{\bf Breakdown points}. The robustness of MCD estimator can be explained by the fact that it has high breakdown points \citep{hampel1971general}.
Specifically, the breakdown point of an estimator measures the smallest fraction of observations that need to be replaced by arbitrary values to carry the estimate beyond all bounds.
Formally,
denote $\mathcal{Y}_{M}$ as a set obtained by replacing $M$ data points of set $\mathcal{Y}$ by some arbitrary values. Then, for a multivariate mean estimator $\mu=\mu(\mathcal Y)$ from $\mathcal Y$, the breakdown point is defined as follows (see Appendix \ref{appendix:preliminaries} for more detailed explanations including the breakdown point of covariance estimator):
\begin{align*}
    \varepsilon^* (\mu, \mathcal{Y})  = \frac{1}{|\mathcal Y|} \min \left\{ M \in \{ 1, \cdots, |\mathcal Y| \} : \sup_{\mathcal{Y}_{M}} \left\| \mu(\mathcal{Y}) -\mu(\mathcal{Y}_{M}) \right\| = \infty \right\}.
\end{align*}
For a multivariate estimator of covariance ${\mathbf \Sigma}$,
we have
\begin{align*}
    \varepsilon^* ({\mathbf \Sigma}, \mathcal{Y})  = \frac{1}{|\mathcal{Y}|} \min \{ M \in \{ 1, \cdots, |\mathcal{Y}| \} : \sup_M \max_i \{ |\log \lambda_i ({\mathbf \Sigma} ( \mathcal{Y})) -\log \lambda_i ({\mathbf \Sigma} ( \mathcal{Y}_{M}))| \}  \},
\end{align*}
where the $k-$th largest eigenvalue of a general $n\times n$ matrix is denoted by $\lambda_k ({\mathbf \Sigma})$, $k=1,\cdots,n$ such that
$\lambda_1 ({\mathbf \Sigma}) \leq \lambda_2 ({\mathbf \Sigma}) \leq \cdots  \leq \lambda_n (\Sigma)$.
This implies that we consider a covariance estimator to be
broken whenever any of the eigenvalues can become
arbitrary large or arbitrary close to 0.

\section{Experimental setup} \label{appendix:setup}

We describe the detailed explanation about all the experiments in Section \ref{sec:experiments}. The code is available at \href{https://anonymous.4open.science/repository/c1980694-d8d7-4b59-94d7-1685cf6aed6a/}{\texttt{[anonymized]}}.

{\bf Detailed model architecture and datasets.}
We consider two state-of-the-art neural network architectures: DenseNet \citep{huang2017densely} and ResNet \citep{he2016deep}.
For DenseNet, our model follows the same setup as in \citet{huang2017densely}: 100 layers, growth rate $k=12$ and dropout rate 0.
Also, we use ResNet with 34 and 44 layers, filters = 64 and dropout rate 0\footnote{ResNet architecture is available at \url{https://github.com/kuangliu/pytorch-cifar}.}.
The softmax classifier is used, and each model is trained by minimizing the cross-entropy loss.
We train DenseNet and ResNet for classifying CIFAR-10 (or 100) and SVHN datasets: the former consists of 50,000 training and 10,000 test
images with 10 (or 100) image classes, and the latter consists of 73,257 training and 26,032 test images with 10 digits.\footnote{We do not use the extra SVHN dataset for training.} 
By following the experimental setup of \citet{ma2018dimensionality},
All networks were trained using SGD with momentum 0.9,
weight decay $10^{-4}$ and an initial learning rate of 0.1. 
The learning rate is divided by 10 after epochs 40 and 80
for CIFAR-10/SVHN (120 epochs in total), and after epochs 80, 120 and 160 for CIFAR-100 (200 epochs in total).
For our method, we extract the hidden features at \{79,89,99\}-th layers and \{27,29,31,33\}-th layers for DenseNet and ResNet, respectively. We assume the uniform class prior distribution.

In the Table \ref{tbl:comparison_with_SOTA_nlp}, we evaluate RoG for the NLP tasks on Twitter and Reuters dataset: the former has a task of part-of-speech (POS) tagging, and the latter has a task of text categorization. The Twitter dataset consists of 14,619 training data from 25 different classes, while some of classes only contain a small number of training data. We exclude such classes that are smaller than 100 in size, then we finally have 14,468 training data and 7,082 test data from 19 different classes. Similarly, we exclude the classes which have a size of less than 100 in the Reuters dataset. Then it consists of 5,444 training data and 2,179 test data from 7 different classes. For training, we use 2-layer FCNs with ReLU non-linearity and uniform noise on both Twitter and Reuters datasets and we extract the hidden features at both layers for 2-layer FCNs. 

\begin{figure} [h] \centering
\subfigure[airplane]
{
\includegraphics[width=0.23\textwidth]{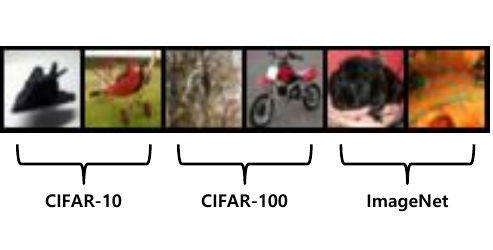}} 
\,
\subfigure[automobile]
{
\includegraphics[width=0.23\textwidth]{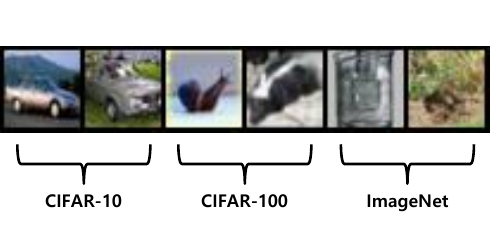}} 
\,
\subfigure[bird]
{
\includegraphics[width=0.23\textwidth]{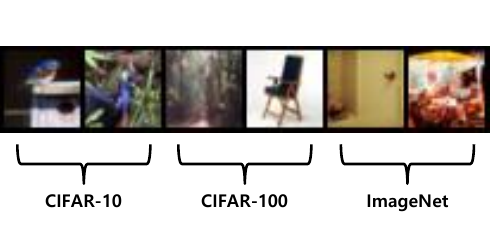}} 
\,
\subfigure[cat]
{
\includegraphics[width=0.23\textwidth]{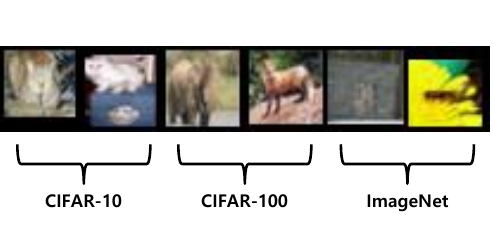}} 
\,
\subfigure[deer]
{
\includegraphics[width=0.23\textwidth]{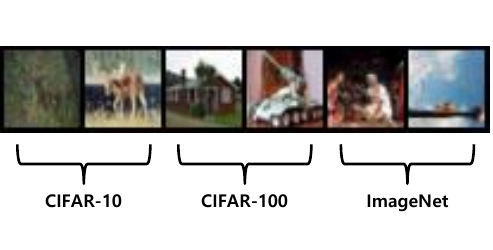}} 
\,
\subfigure[dog]
{
\includegraphics[width=0.23\textwidth]{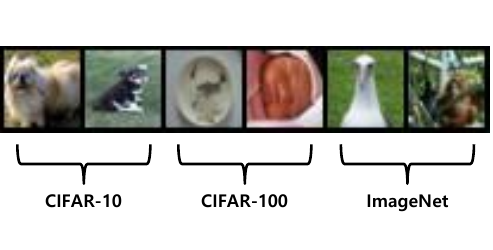}} 
\,
\subfigure[frog]
{
\includegraphics[width=0.23\textwidth]{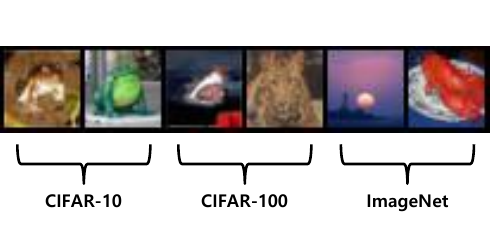}} 
\,
\subfigure[horse]
{
\includegraphics[width=0.23\textwidth]{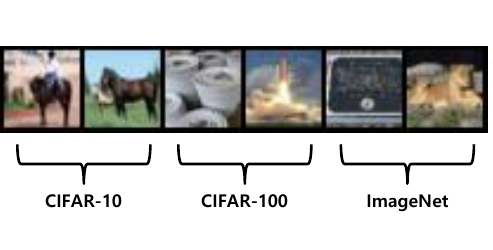}} 
\,
\subfigure[ship]
{
\includegraphics[width=0.23\textwidth]{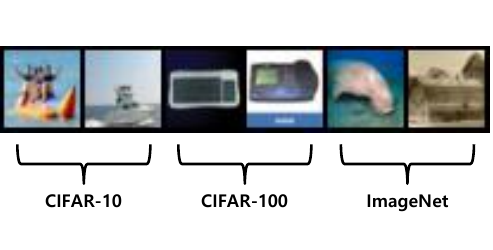}} 
\,
\subfigure[truck]
{
\includegraphics[width=0.23\textwidth]{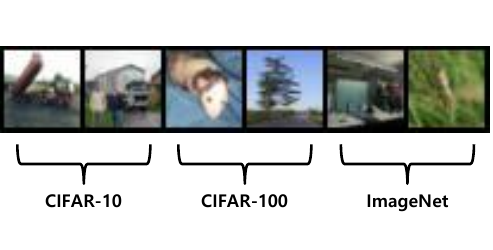}} 
\caption{Examples of open-set noise in CIFAR-10 dataset.} \label{fig:open_Set_total}
\end{figure}

We also consider the open-set noisy scenarios \citep{wang2018iterative}. 
Figure \ref{fig:open_Set_total} shows the examples of open-set noisy dataset which is built by replacing some training images in CIFAR-10 by external images in CIFAR-100 and Downsampled ImageNet \citep{chrabaszcz2017downsampled} which is equivalent to the ILSVRC 1,000-class ImageNet dataset \citep{deng2009imagenet}, but with images downsampled to $32 \times 32$ resolution. 
In the Table \ref{tbl:openset_noise}, 
we maintain the original label of the CIFAR-10 dataset, while replacing 60\% of the training data in CIFAR-10 with training data in CIFAR-100 and Downsampled ImageNet.

{\bf Validation}. For our methods, the ensemble weights are chosen by optimizing the NLL loss over the validation set. We assume that the validation set consists of 1000 images with (same type and fraction of) noisy labels. However, one can expect that if we use all validation samples, the performance of our method can be affected by outliers. To relax this issue, we use only half of them, chosen by the MCD estimator.
Specifically, we first compute the Mahalanobis distance for all validation samples using the parameters from MCD estimator, and select 500 samples with smallest distance. Then, one can expect that our ensemble method is more robust against the noisy labels in validation sets. In the case of Twitter and Reuters dataset, validation set consists of 570 and 210 samples with noisy labels, respectively.

{\bf Training method for noisy label learning}.
We consider the following training methods for noisy label learning:
\begin{itemize}
    \item [(a)] {\bf Hard bootstrapping} \citep{reed2014training}: Training with new labels generated by a convex combination (the “hard” version) of the noisy labels and their predicted labels.
    \item[(b)] {\bf Soft bootstrapping} \citep{reed2014training}: Training with new labels generated by a convex combination (the “soft” version) of the noisy labels and their predictions.
    \item[(c)] {\bf Backward} \citep{patrini2017making}: Training via loss correction by multiplying the cross-entropy loss by a noise-aware correction matrix.
    \item[(d)] {\bf Forward} \citep{patrini2017making}: Training with label correction by multiplying the network prediction by a noise-aware correction matrix.
    \item[(e)] {\bf Forward (Gold)} \citep{hendrycks2018using}: An augmented version of Forward method which replaces its corruption matrix estimation with the identity on trusted samples.
    \item[(f)] {\bf GLC (Gold Loss Correction)} \citep{hendrycks2018using}: Training with the corruption matrix which is estimated by using the trusted dataset.
    \item[(g)]{\bf D2L} \citep{ma2018dimensionality}: Training with new labels generated by a convex combination of the noisy labels and their predictions, where its weights are chosen by utilizing the Local Intrinsic Dimensionality (LID).
    \item[(h)] {{\bf Decoupling} \citep{malach2017decoupling}: Updating the parameters only using the samples which have different prediction from two classifier.}
    \item[(i)] {{\bf MentorNet} \citep{jiang2017mentornet}: An extra teacher network is pre-trained and then used to select clean samples for its student network.}
    \item[(j)] {{\bf Co-teaching} \citep{han2018co}: A simple ensemble method where each network selects its small-loss training data and back propagates the training data selected by its peer network.}
    \item[(k)] {\bf Cross-entropy}: the conventional approach of training with cross-entropy loss.
\end{itemize}

\section{Layer-wise characteristics of generative classifiers} \label{appendix:layer_wise}

\begin{figure} [h] \centering
\subfigure[Generalization from noisy labels]
{
\includegraphics[width=0.32\textwidth]{ResNet34_layer_wise_Noise.pdf}} 
\,
\subfigure[Generalization from noisy labels]
{
\includegraphics[width=0.32\textwidth]{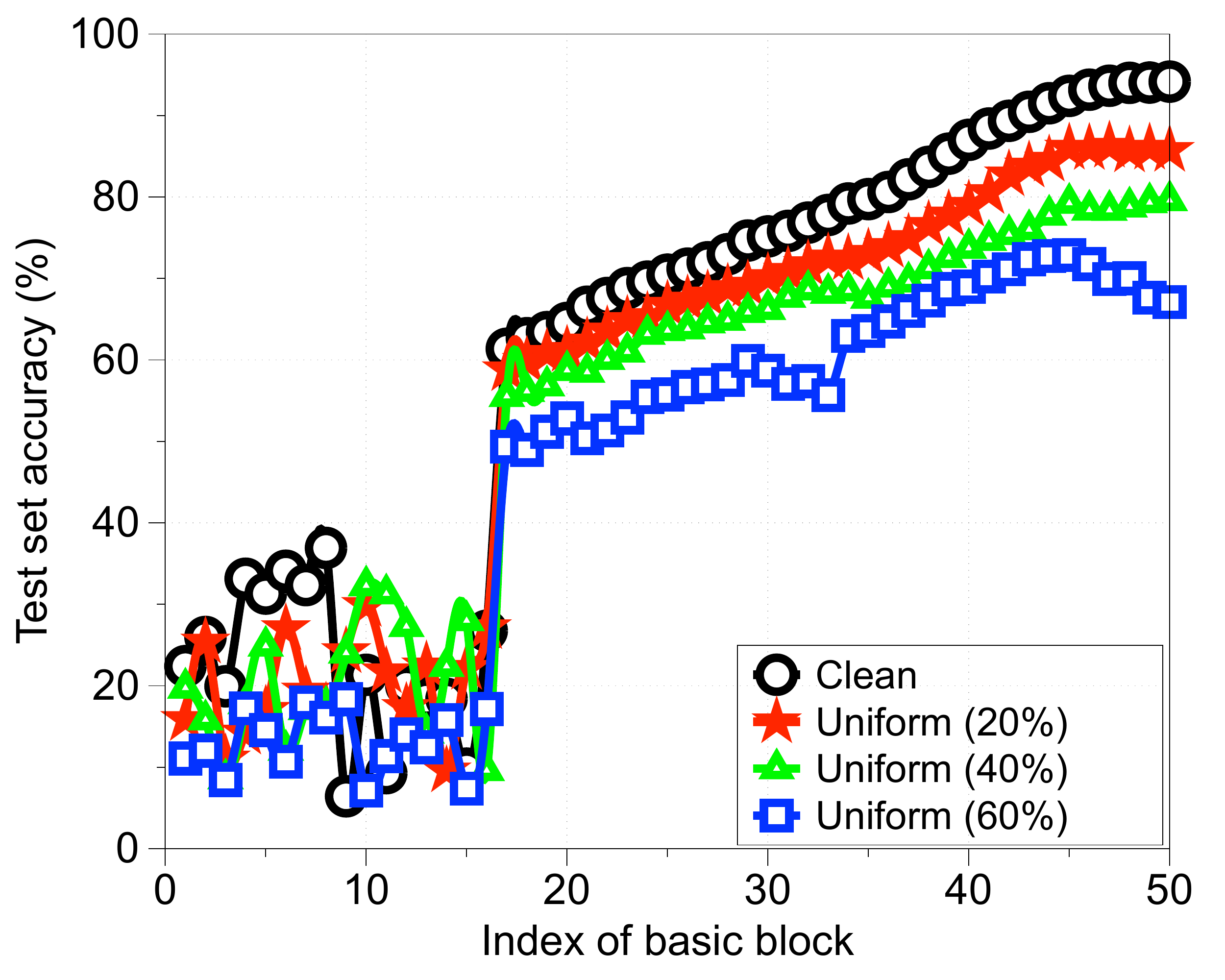}}
\caption{Layer-wise characteristics of generative classifiers from (a) ResNet-34 and (b) DenseNet-100 trained on the CIFAR-10 dataset.} \label{fig:layer_wise_appendx}
\end{figure}

Figure~\ref{fig:layer_wise_appendx} shows the classification accuracy of the generative classifiers from different basic blocks of ResNet-34 \citep{he2016deep} and DenseNet-100 \citep{huang2017densely}.
One can note that the generative classifiers from DenseNet and ResNet have different patterns due to the architecture design.
In the case of DenseNet, we found that it produces meaningful features after 20-th basic blocks. 

\section{{Proof of Theorem \ref{th:main}}} \label{appendix:proof} 

In this section, we present a proof of Theorem \ref{th:main}, {which consists of two statements: the limit of estimation error (\ref{ineq1:mahalanobis}) and estimated error ratio (\ref{ineq2:mahalanobis}). We prove both statements one by one as stated in below. For convenience, we skip to mention the Continuous Mapping Theorem\footnote{P. Billingsley, Convergence of Probability Measures, John Wiley \& Sons, 1999} and} the number of training samples $N$ goes to infinity for all convergences in the proof. 

\subsection{Proof of the limit of estimation error (\ref{ineq1:mahalanobis})}

We start with a following lemma, which shows 
{the convergences} of sample and MCD estimators 
as the number of training samples $N$ goes to infinity. 

\begin{lemma} \label{lem:main}
Suppose we have $N$ number of $d$-dimensional training samples {$\mathcal{X}_N=\{\mathbf{x}_1, \cdots, \mathbf{x}_N\}$ and $\mathcal{X}_N$ contains outlier samples with the fixed fraction $\delta_{\tt out}<1$. We assume the outlier samples are from an arbitrary distribution $P_{\tt out}$} with zero mean and finite covariance matrix $\sigma_{\tt out}^2 {\mathbf I}$, and the clean samples are from a distribution of the hidden features of DNNs $P_{\tt clean}$ with mean $\mathbf{\mu}$ and covariance matrix $\sigma^2 {\mathbf I}$. 
Let $\Bar{\mu}$ and $\Bar{\mathbf{\Sigma}}$ be the mean and covariance matrix of sample estimator, and let $\widehat \mu$ and $\widehat{\mathbf{\Sigma}}$ be the mean and covariance matrix of MCD estimator which selects {samples from $\mathcal{X}_N$ with the fixed fraction $\frac{d}{N}<\delta_{\tt mcd}<1$} to optimize its objective (\ref{eq:mcdopt}). Then the {mean and covariance matrix of} sample estimator converge {almost surely to below} as $N\rightarrow\infty:$
\begin{align*}
    &\Bar{\mu} \overset{\tt{a.s.}}{\rightarrow} \left( 1-\delta_{\tt out} \right)\mu,\quad  \Bar{\mathbf{\Sigma}}\overset{\tt{a.s.}}{\rightarrow} \left(\left( 1-\delta_{\tt out} \right)\sigma^2 + \delta_{\tt out} \sigma_{\tt out}^2\right)\mathbf{I} + \delta_{\tt out} \left(1-\delta_{\tt out} \right)\mu\mu^T.
\end{align*}
{In addition}, if $\delta_{\tt mcd}\le 1-\delta_{\tt out}$ and {$\sigma^2<\sigma_{\tt out}^2$, the mean and covariance matrix of} MCD estimator converge {almost surely} to below as $N\rightarrow\infty:$
\begin{align*}
    \widehat\mu \overset{\tt{a.s.}}{\rightarrow}\mu, \qquad\mathbf{\widehat{\Sigma}} \overset{\tt{a.s.}}{\rightarrow} \sigma^2\mathbf{I}.
\end{align*}
\end{lemma}

A proof of the lemma is given in appendix \ref{appendix:lemma_proof}, {where it is built upon the fact that the determinant of covariance matrix with some assumptions can be expressed as the $d$-th degree polynomial of outlier ratio.}

Lemma \ref{lem:main} states the convergences of sample and MCD estimators on a single distribution of hidden features of DNNs. Without loss of generality, one can assume the mean of outlier distribution is zero, i.e., $\mu_{\tt out} = 0$ by an affine translation of hidden features. Furthermore, 
one can extend Lemma \ref{lem:main} to $C$ number of distributions, which have the class mean $\mu_c$ and class covariance matrix $\mathbf{\Sigma}_c$ on each class label $c\in \{1,...,C\}$ with the assumptions $\mathcal A1 \sim \mathcal A4$. Then the class mean of MCD and sample estimators converge almost surely as follows: 
\begin{align*}
    \widehat \mu_c \overset{\tt{a.s.}}{\rightarrow}\mu_c, \qquad 
    \Bar \mu_c \overset{\tt{a.s.}}{\rightarrow}\left(1-\delta_{\tt out}\right)\mu_c,
\end{align*}
which implies that
\begin{align*}
    \|\mu_c - {\widehat \mu}_c\|_1 \overset{\tt{a.s.}}{\rightarrow} 0,
    \qquad \|\mu_c - {\Bar \mu}_c\|_1 \overset{\tt{a.s.}}{\rightarrow} \delta_{\tt out}\|\mu_c\|_1.
\end{align*}
This completes the proof of the limit of estimation error (\ref{ineq1:mahalanobis}).

\subsection{Proof of the limit of estimated error ratio (\ref{ineq2:mahalanobis})}

Recall the class mean of MCD and sample estimators converge almost surely as follows:
\begin{align*}
    \widehat \mu_c \overset{\tt{a.s.}}{\rightarrow}\mu_c, \qquad 
    \Bar \mu_c \overset{\tt{a.s.}}{\rightarrow}\left(1-\delta_{\tt out}\right)\mu_c.
\end{align*}

Then one can induce that the limit of mean distance of sample and MCD estimators as follow:
\begin{align}
    &\|\Bar{\mu}_c - \Bar{\mu}_{c^\prime}\|_2 \overset{\tt{a.s.}}{\rightarrow} 
    \left(1 - \delta_{\tt out}\right)^2\|{\mu}_i - {\mu}_c\|_2,\label{sample_mean_distance}\\
    &\|\widehat{\mu}_c - \widehat{\mu}_{c^\prime}\|_2 \overset{\tt{a.s.}}{\rightarrow} 
    \|{\mu}_c - {\mu}_{c^\prime}\|_2.\label{mcd_mean_distance}
\end{align}

On the other hand, the assumptions $\mathcal A1$ states that all class covariance matrices are the same, i.e., $\mathbf{\Sigma}_c=\sigma^2 {\mathbf I}$. 
Then tied covariance matrices $\mathbf{\Bar{\Sigma}}$ and $\mathbf{\widehat{\Sigma}}$ are given by gathering $\mathbf{\Bar{\Sigma}}_c$ and $\mathbf{\widehat{\Sigma}}_c$ on each class $c$ respectively:
\begin{align}
    \mathbf{\Bar{\Sigma}} = \frac{\sum_c N_c \mathbf{\Bar{\Sigma}}_c}{\sum_c N_c} = \frac{\sum_c \mathbf{\Bar{\Sigma}}_c}{C}, 
    \quad 
    \mathbf{\widehat{\Sigma}} = \frac{\sum_c K_c \mathbf{\widehat{\Sigma}}_c}{\sum_c K_c} = \frac{\sum_c \mathbf{\widehat{\Sigma}}_c}{C}. \label{eq:tied_cov_induced}
\end{align}
From the tied covariance matrices (\ref{eq:tied_cov_induced}) and Lemma \ref{lem:main}, one can induce their convergences and limits as follow:
\begin{align}
    &\mathbf{\Bar{\Sigma}} \overset{\tt{a.s.}}{\rightarrow} 
    \left(\left(1-\delta_{\tt out}\right)\sigma^2 + \delta_{\tt out}\sigma_{\tt out}^2\right)\mathbf{I} + \delta_{\tt out}\left(1-\delta_{\tt out}\right)\frac{1}{C}\sum_c \mu_c\mu_c^T,\label{sample_cov_limit} \\
    &\mathbf{\widehat{\Sigma}} \overset{\tt{a.s.}}{\rightarrow} 
    \sigma^2\mathbf{I}.\label{mcd_cov_limit}
\end{align}

Next, we define a function of the tied covariance matrix as follow:
\begin{align*}
    \phi({\mathbf {\widehat \Sigma}}) = {4{\|\mathbf {\widehat \Sigma}^{-1}\|_2}{\|\mathbf {\widehat \Sigma}\|_2}}
    {\left(1+{\|\mathbf {\widehat \Sigma}^{-1}\|_2}{\|\mathbf {\widehat \Sigma}\|_2}\right)^{-2}}.
\end{align*}
Then (\ref{mcd_cov_limit}) implies that $\phi({\mathbf {\widehat \Sigma}}) \overset{\tt{a.s.}}{\rightarrow} 1$ clearly. 
Since the condition number $t$ of $\mathbf {\widehat \Sigma}^{-1}$ is in $[1,\infty)$, i.e. $t = \|\mathbf {\widehat \Sigma}^{-1}\|_2 \|\mathbf {\widehat \Sigma}\|_2 \in [1,\infty)$,
one can induces that $\phi({\mathbf {\widehat \Sigma}}) = \phi(t) = \frac{4t}{(1+t)^2}$ for $t\in [1,\infty)$ and it is a monotonic decreasing function by using the change of variables with $t$. Hence $\phi(t)$ has the maximum at $t=1$ and it implies
\begin{align}
    1 = \phi(1) =
    \lim_{N\rightarrow\infty}{\phi({\mathbf {\widehat \Sigma}})} \geq
    \lim_{N\rightarrow\infty} {\phi({\mathbf {\Bar \Sigma}})}. \label{cond_ratio}
\end{align}
Therefore the limits of mean distance of estimators (\ref{sample_mean_distance}), (\ref{mcd_mean_distance}) and ratio of the function $\phi$ (\ref{cond_ratio}) hold the statement (\ref{ineq2:mahalanobis}),
\begin{align*}
    {\frac{\phi({\mathbf {\widehat \Sigma}})\|\widehat{\mu}_c - \widehat{\mu}_{c^\prime}\|_2}{\phi({\mathbf {\Bar \Sigma}})\|\Bar{\mu}_c - \Bar{\mu}_{c^\prime}\|_2}\overset{\tt a.s.}{\rightarrow}
    \lim_{N\rightarrow\infty}}\frac{\phi({\mathbf {\widehat \Sigma}})\|\widehat{\mu}_c - \widehat{\mu}_{c^\prime}\|_2}{\phi({\mathbf {\Bar \Sigma}})\|\Bar{\mu}_c - \Bar{\mu}_{c^\prime}\|_2} =
    \lim_{N\rightarrow\infty} \frac{1}{\left(1-\delta_{\tt out}\right)^2 \phi({\mathbf {\Bar \Sigma}})}  \geq 1.
\end{align*}

This completes the proof of Theorem \ref{th:main}.

\subsection{{Proof of Lemma \ref{lem:main}}} \label{appendix:lemma_proof} 

In this part, we present a proof of Lemma \ref{lem:main}. We show the almost surely convergences of sample and MCD estimators as the number of training samples $N$ goes to infinity. 

{\bf Proof of the convergence of sample estimator}. 
First of all, the set of training samples $\mathcal{X}_N=\{\mathbf{x}_1, \cdots, \mathbf{x}_N\}$ contains outlier samples with the fixed fraction $\delta_{\tt out}$. So, $\mathcal{X}_N$ is from a mixture distribution $P_{{\tt mix}}=\left( 1-\delta_{\tt out} \right)P_{\tt clean} + \delta_{\tt out} P_{\tt out}$. Then mean and covWariance matrix of sample estimator, $\Bar{\mu}$ and $\Bar{\mathbf{\Sigma}}$, estimate mean $\mu_{\tt mix}$ and covariance matrix $\mathbf{\Sigma}_{\tt mix}$ of the mixture distribution $P_{\tt mix}$, respectively. One can induce $\mu_{\tt mix}$ and $\mathbf{\Sigma}_{\tt mix}$ directly as follow:
\begin{align}
    \mu_{\tt mix}=\left( 1-\delta_{\tt out} \right)\mu, \quad \mathbf{\Sigma}_{\tt mix} = \left( 1-\delta_{\tt out} \right)\sigma^2\mathbf{I} + \delta_{\tt out} \sigma_{\tt out}^2\mathbf{I} + \delta_{\tt out}\left( 1-\delta_{\tt out} \right)\mu\mu^T.\label{eq:mixtrue_distribution}
\end{align}

Since $P_{\tt mix}$ has the finite covariance matrix, i.e., $\mathbf{\Sigma}_{\tt mix}<\infty$, one can apply the the Strong Law of Large Numbers\footnote{W. Feller, An Introduction to Probability Theory and Its Applications, John Wiley \& Sons, 1968} to the sample estimator of the mixture distribution $P_{\tt mix}$. Then the mean and covariance matrix of sample estimator converge almost surely to the mean and covariance matrix of $P_{\tt mix}$, respectively:
\begin{align*}
    \Bar{\mu} \overset{\tt{a.s.}}{\rightarrow} \mu_{\tt mix}, \qquad 
    \Bar{\mathbf{\Sigma}} \overset{\tt{a.s.}}{\rightarrow} \mathbf{\Sigma}_{\tt mix}.
\end{align*}
This completes the proof of the convergence of sample estimator. 

{\bf Proof of the convergence of MCD estimator}.
Consider a collection $E_{\tt q}$ of subsets $\mathcal{X}_{K,q}\subset \mathcal{X}_N$ with the size $K\left(=\lfloor \delta_{\tt mcd}N \rfloor\right)$, and each subset $\mathcal{X}_{K,q} \in E_{\tt q}$ contains the outlier samples with the fraction $q\in[0, 1]$. Then $\mathcal{X}_{K,q}\in E_{\tt q}$ is from a mixture distribution $P_{\tt q}=\left( 1-q \right)P_{\tt clean} + q P_{\tt out}$. One can induce that the mean $\mu_{\tt q}$ and covariance matrix $\mathbf{\Sigma}_{\tt q}$ of the mixture distribution $P_{\tt q}$ as (\ref{eq:mixtrue_distribution}):
\begin{align}
    {\mu_{\tt q}=(1-q)\mu, \quad 
    \mathbf{\Sigma}_{\tt q} = (1-q)\sigma^2\mathbf{I} + q\sigma_{\tt out}^2\mathbf{I} + q(1-q)\mu\mu^T.\label{eq:mixtrue_distribution2}}
\end{align}
{Thus sample mean estimator $\Bar{\mu}_{\mathcal{X}_{K,q}}$ and covariance estimator $\mathbf{\Bar{\Sigma}}_{\mathcal{X}_{K,q}}$ of a subset ${\mathcal{X}_{K,q}}$ converge almost surely to $\mu_{\tt q}$ and $\mathbf{\Sigma}_{\tt q}$ respectively:
\begin{align*}
    \Bar{\mu}_{\mathcal{X}_{K,q}} \overset{\tt{a.s.}}{\rightarrow} \mu_{\tt q}, \qquad
    \mathbf{\Bar{\Sigma}}_{\mathcal{X}_{K,q}} \overset{\tt{a.s.}}{\rightarrow} \mathbf{\Sigma}_{\tt q},
\end{align*}
by the Strong Law of Large Numbers.}

On the other hand, there is a subset $\mathcal{X}_{K,q^*}^*\subset \mathcal{X}_N$ in $E_{{\tt q}^*}$ which is selected by MCD estimator. Then the determinant of its covariance matrix is the minimum over all subset of size $K$ in $\mathcal{X}_N$, and 
$\Bar{\mu}_{\mathcal{X}_{K,q^*}^*}=\widehat{\mu} \overset{\tt{a.s.}}{\rightarrow} \mu_{\tt {q^*}}$,  $\mathbf{\Bar{\Sigma}}_{\mathcal{X}_{K,q^*}^*}=\widehat{\mathbf{\Sigma}} \overset{\tt{a.s.}}{\rightarrow} \mathbf{\Sigma}_{\tt {q^*}}$ as $N\rightarrow\infty$. Since the determinant is a continuous funciton, the Continuous Mapping Theorem\footnote{P. Billingsley, Convergence of Probability Measures, John Wiley \& Sons, 1999} implies 
\begin{align*}
    \underset{\mathcal{X}_{K,q} \in E_{\tt q}, \forall q}{\min} \det (\mathbf{\Bar{\Sigma}}_{\mathcal{X}_{K,q}})
    \overset{\tt{a.s.}}{\rightarrow} \underset{q}{\min}\:
    \det\left(\mathbf{\Sigma}_{\tt {q}}\right),
\end{align*}
{and }
\begin{align*}
    \underset{\mathcal{X}_{K,q} \in E_{\tt q}, \forall q}{\min} \det (\mathbf{\Bar{\Sigma}}_{\mathcal{X}_{K,q}}) = \det (\mathbf{\Bar{\Sigma}}_{\mathcal{X}_{K,q^*}^*}) = 
    \det (\mathbf{\widehat{\Sigma}}) \overset{\tt{a.s.}}{\rightarrow} \det\left(\mathbf{\Sigma}_{\tt {q^*}}\right).
\end{align*}

{Now, we'd like to show}
\begin{align}
    {\underset{q}{\min}\:\det\left(\mathbf{\Sigma}_{\tt {q}}\right) = 
    \det\left(\mathbf{\Sigma}_{\tt {q^*}}\right) =
    \det\left(\mathbf{\Sigma}_{\tt {0}}\right),\label{eq:lemma_objective}}
\end{align}
to complete the proof of Lemma \ref{lem:main}. 

By the assumption $\delta_{\tt mcd}\le 1-\delta_{\tt out}$, $E_{\tt 0}$ is non-empty. It shows the existence of $\mathbf{\Sigma}_{\tt {0}}$. 
From the covariance matrix $\mathbf{\Sigma}_{\tt {q}}$ (\ref{eq:mixtrue_distribution2}),  $\det(\mathbf{\Sigma}_{\tt {q}})$ is a $d$-th degree polynomial of $q$ as follow:
\begin{align*}
    {\det(\mathbf{\Sigma}_{\tt {q}})}&{~= 
    \mathrm{det}\left((1-q)\sigma^2\mathbf{I} + q\sigma_{\tt out}^2\mathbf{I} + q(1-q)\mu\mu^T\right)}\\
    &{~= \left((1-q)\sigma^2 + q\sigma_{\tt out}^2\right)^{d-1}\left((1-q)\sigma^2 + q\sigma_{\tt out}^2 + q(1-q)\mu^T\mu\right).}
\end{align*}
{Since the assumption gives $\sigma_{\tt out}^2>\sigma^2$, ~$\det(\mathbf{\Sigma}_{\tt {q}})$ has the lower bound $\det(\mathbf{\Sigma}_{\tt {0}})$ as follow:}
\begin{align*}
    {\det(\mathbf{\Sigma}_{\tt {q}})}&{~=\left((1-q)\sigma^2 + q\sigma_{\tt out}^2\right)^{d-1}\left((1-q)\sigma^2 + q\sigma_{\tt out}^2 + q(1-q)\mu^T\mu\right)}\\
    &{~\ge \left(\sigma^2\right)^{d-1}\left(\sigma^2 + q(1-q)\mu^T\mu\right)}\\
    &{~\ge \left(\sigma^2\right)^{d-1}\sigma^2 = \det(\mathbf{\Sigma}_{\tt {0}}).}
\end{align*}
Then $\det(\mathbf{\Sigma}_{\tt {q}})\ge\det(\mathbf{\Sigma}_{\tt {0}})$ for all  $q\in[0,1]$ and the equality holds for only $q=0$. It implies $q^*=0$ and (\ref{eq:lemma_objective}) is the shown. Therefore the mean and covariance matrix of MCD estimator converge almost surely to $\mu$ and $\sigma^2\mathbf{I}$
, respectively:
\begin{align*}
    \widehat \mu\overset{\tt{a.s.}}{\rightarrow} \mu_{\tt 0}=\mu, \quad \mathbf{\widehat{\Sigma}}\overset{\tt{a.s.}}{\rightarrow}\mathbf{\Sigma}_{\tt {0}}=\sigma^2\mathbf{I}.
\end{align*}
This completes the proof of Lemma \ref{lem:main}.

\section{Comparison with robust clustering methods} \label{app:cluster}

To remove the outliers (i.e., samples with noisy labels) in hidden feature, one can consider other robust clustering methods to estimate the parameters of generative classifiers. In this section, we test trimmed K-means (TKM) \citep{garcia1999robustness} as a new baseline, and compare the performances with MCD estimator.
Specifically, we use noisy labels to initialize clusters of TKM, and assign a "majority" label to each trained cluster. 
For pair comparison, a tied covariance across clusters is assumed, and we run the same number of iterations for both TKM and MCD.
Table \ref{tb8:TKM} reports the corresponding results under ResNet and CIFAR-10 with uniform noises when we induce a generative classifier only using a penultimate layer.
First, we remark that TKM outperforms Softmax, which supports our claim that the clustering property of DNN features is useful to handle noisy labels.
However, the generative classifier with MCD estimator still outperforms all baselines because it utilizes the information of noisy labels carefully to derive a new decision rule at all iterations, while TKM is essentially an unsupervised method and does not utilize the information, except for initialization and termination.

\begin{table}[t]
\centering
\begin{tabular}{c|c|cccccccccccc}
\toprule
 \multirow{2}{*}{Model} &\multirow{2}{*}{Inference method} & \multicolumn{4}{c}{CIFAR-10} \\
&     & Clean & \multicolumn{1}{l}{Uniform (20\%)} & Uniform (40\%) & Uniform (60\%) \\ \midrule
\multirow{3}{*}{ResNet} 
&Softmax
& 94.76\%  & \multicolumn{1}{c}{80.88\%}  
& 61.98\% & {39.96\%}  
\\
& Generative + TKM
& 94.35\% & \multicolumn{1}{c}{81.41\%}  
& 63.27\% & {41.78\%}  
\\
& Generative + MCD (ours)
& 94.76\% & \multicolumn{1}{c} {\bf{83.86}\%}  
&\bf{68.03}\% & \bf{44.87}\%
\\\bottomrule
\end{tabular}
\caption{
Test accuracy (\%) of ResNet on the CIFAR-10 dataset with uniform noise. The best results are highlighted in bold if the gain is bigger than 1\%.}\label{tb8:TKM}
\end{table}

\end{document}